\documentclass[10pt,journal,cspaper,compsoc]{IEEEtran}

\usepackage{graphicx}
\usepackage{amsmath,amssymb}
\usepackage{times}
\usepackage{multirow}
\usepackage{ragged2e}
\usepackage{multirow,array}
\usepackage{cite}
\usepackage{rotating}
\usepackage{multicol,multirow}
\usepackage{overpic}
\usepackage{contour}
\usepackage{booktabs}       
\usepackage{epsfig}
\usepackage{graphicx}
\usepackage{soul}

\usepackage{amsmath,amsfonts}
\usepackage{algorithmic}
\usepackage{algorithm}
\usepackage{array}
\usepackage[caption=false,font=normalsize,labelfont=sf,textfont=sf]{subfig}
\usepackage{textcomp}
\usepackage{booktabs}
\usepackage{stfloats}
\usepackage{url}
\usepackage{verbatim}
\usepackage{graphicx}
\usepackage{cite}
\usepackage{times}
\usepackage{epsfig}
\usepackage{amsmath}
\usepackage{amssymb}
\usepackage{color}
\usepackage{multicol, blindtext}
\usepackage{multirow}
\usepackage[dvipsnames]{xcolor}
\usepackage{cancel}
\usepackage{bm}
\usepackage{gensymb}

\usepackage[colorlinks,bookmarksnumbered,bookmarksopen,citecolor=black]{hyperref}
\usepackage{silence}
\usepackage{pifont}

\def\ie{\emph{i.e.}}

\definecolor{mypurple}{RGB}{160, 48, 160}
\definecolor{mygreen}{RGB}{84, 130, 53}
\definecolor{myorange}{RGB}{255, 210, 0}
\definecolor{myblue}{RGB}{0, 0, 255}
\definecolor{mybrown}{RGB}{190, 62, 19}
\newcommand{\Rows}[1]{\multirow{2}{*}{#1}}

\newcommand{\bl}[1]{\textcolor{myblue}{#1}}
\newcommand{\re}[1]{\textcolor{red}{#1}}
\renewcommand{\paragraph}[1]{\vspace{3mm} \noindent \textbf{#1}}

\begin{document}

\title{What Images are More Memorable to Machines?}

\author{Junlin Han, Huangying Zhan, Jie Hong, Pengfei Fang, Hongdong Li, Lars Petersson, Ian Reid
\IEEEcompsocitemizethanks{
  \IEEEcompsocthanksitem Junlin Han, Jie Hong, Pengfei Fang, and Hongdong Li are with the Australian National University (ANU), Canberra, ACT, 2601, Australia;
  \IEEEcompsocthanksitem Huangying Zhan and Ian Reid are with the University of Adelaide, Adelaide, SA, 5000, Australia;
 \IEEEcompsocthanksitem Lars Petersson is with Commonwealth Scientific and Industrial Research Organisation (CSIRO), Canberra, ACT, 2601, Australia.
}
}

\markboth{IEEE Transactions on Pattern Analysis and Machine Intelligence Submission}%
{Bian \MakeLowercase{\textit{et al.}}: What Images are More Memorable to Machines}


\IEEEtitleabstractindextext{%
\begin{abstract}
\justifying
This paper studies the problem of measuring and predicting how memorable an image is to pattern recognition machines, as a path to explore machine intelligence. Firstly, we propose a self-supervised machine memory quantification pipeline, dubbed ``MachineMem measurer'', to collect machine memorability scores of images. Similar to humans, machines also tend to memorize certain kinds of images, whereas the types of images that machines and humans memorize are different. Through in-depth analysis and comprehensive visualizations, we gradually unveil that``complex" images are usually more memorable to machines. We further conduct extensive experiments across 11 different machines (from linear classifiers to modern ViTs) and 9 pre-training methods to analyze and understand machine memory. This work proposes the concept of machine memorability and opens a new research direction at the interface between machine memory and visual data. 
\end{abstract}

}
\maketitle
\IEEEraisesectionheading{\section{Introduction}\label{sec:intro}}


\IEEEPARstart{T}he philosophical question posed by Alan Turing in 1950 -- "Can machines think?"\cite{alan1950} has underpinned much of the pursuit within the realm of artificial intelligence (AI). Despite the significant strides made in the last few decades, an unambiguous answer to this question remains elusive. Noteworthy developments in computer vision~\cite{alexnet,vgg,resnet,dosovitskiy2020image} and artificial intelligence~\cite{radford2018gpt,devlin2018bert,vaswani2017attention,silver2017mastering} have indeed redefined our capabilities in domains from vision-driven autonomous agents~\cite{geiger2012we,he2017mask} to multiple inter-disciplinary fields~\cite{jumper2021highly,ravuri2021skilful,akiyama2019first}.

The power of pattern recognition machines today is staggering. But due to the nature of back-propagation~\cite{rumelhart1985learning} and black box models~\cite{guidotti2018survey}, decisions made by machines can sometimes be untrustworthy. Therefore, a deeper comprehension of machine intelligence is a vital step towards crafting machines that are not just powerful, but even more reliable and understandable. In addition, discerning the similarities and differences between artificial and natural intelligence could lay the foundation for the creation of machines capable of perceiving, learning, and thinking in a manner more akin to humans~\cite{lake2017building}. In this paper, we propose an exploration of machine intelligence from the perspective of memory, a pivotal component in both intelligence and cognition~\cite{colom2010human}. 

When we are looking at images, we humans can naturally perceive the same types of information, and thus tend to behave similarly in memorizing images~\cite{lamem}. That is, some images sharing certain patterns are more memorable to all of us~\cite{isola2013makes,lamem,goetschalckx2019ganalyze}. Recalling our proposal on exploring machine intelligence from a view of memory, we dive into: How well do machines memorize certain types of images? What attributes make an image memorable to a machine? Do different machines exhibit varying memory characteristics?

Our exploration begins with a quantification of machine memory. We adopt the human memorability score (HumanMem score)~\cite{isola2011makes} concept and propose the machine memorability score (henceforth referred to as MachineMem score) as a measure of machine memory. Our next goal is to collect MachineMem scores for a variety of images. To accomplish this, we introduce a novel framework, the MachineMem measurer, inspired by the repeat detection task and visual memory game~\cite{isola2013makes}. This framework produces MachineMem scores for images in a self-supervised manner, allowing us to label the entire LaMem dataset~\cite{lamem}. Based on this, we train a regression model to predict MachineMem scores in real-time and introduce some advanced training techniques to enhance the performance of both human and machine memorability score predictors.

\begin{figure*}[h!]
\begin{minipage}{\textwidth}
	\centering
	\vspace{0.0in}
     \includegraphics[width = \textwidth]
     {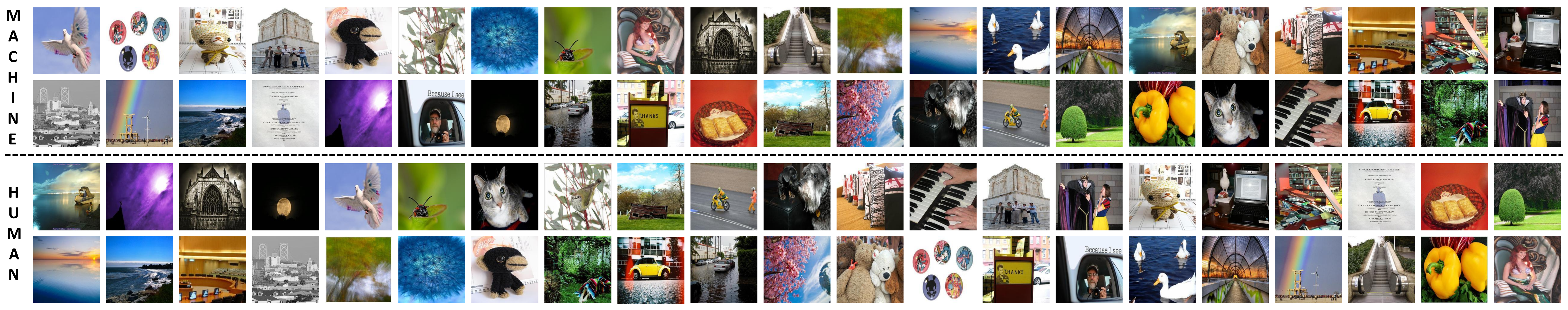}
     \caption{\textbf{Sample images} that are sorted from less memorable (left) to more memorable (right) for both machines and humans. The first top two rows are arranged by machine memorability scores and the bottom two rows are sorted by human memorability scores. 
     }
     \label{fig:lowtohigh}
\end{minipage}
\end{figure*}

Armed with collected MachineMem scores, we delve into the investigation of what makes an image memorable to machines. This exploration involves multiple approaches, beginning with a visual teaser (Figure~\ref{fig:lowtohigh}), wherein sample images are arranged by their MachineMem scores. A quantitative analysis follows, presenting the correlations between MachineMem scores and 13 image attributes. We then analyze the memorability of different classes and scenes for machines. Further, we apply GANalyze~\cite{goetschalckx2019ganalyze} to visualize how the memorability of a particular image changes with varying MachineMem scores. A comparative analysis between machine memory and human memory helps highlight the differences and similarities. Lastly, we study machine memorability of images for different types of machines.

We further aim to understand machine memorability more deeply. We conduct two case studies that analyze the MachineMem scores produced by 11 different machines (ranging from conventional machines to various modern DNNs) and varying pretext tasks (pre-training models). The final part of our study investigates the consistency of machine memorability across different training settings.

\section{Related Work}
\label{sec:related}

\paragraph{Visual cognition and memory of humans.}
Pioneering studies~\cite{isola2011makes,isola2013makes} have systematically explored the elements that make a generic image memorable to humans. They established a visual memory game, a repeat detection task that runs through a long stream of images. This game involves multiple participants, and the averaged accuracy of detecting repeated images provides a quantified HumanMem score for each image. Subsequent research in this area~\cite{lahrache2022survey,zhang2020review,bylinskii2022memorability} has created more datasets~\cite{lamem,bainbridge2013intrinsic,lu2020understanding,goetschalckx2019memcat,dubey2015makes} and developed more powerful methods for predicting HumanMem scores~\cite{lamem,kim2013relative,celikkale2013visual,peng2015predicting,fajtl2018amnet,perera2019image,lu2020understanding,jaegle2019population,leyva2021video}. One of the goals of this work is to compare machine memory and human memory. To this end, we preserve the definition of MachineMem score, mirroring the HumanMem score, and incorporate the key design elements of the visual memory game into our MachineMem measurer.

In psychology and cognition research, memory is broadly divided into sensory~\cite{pearson2008sensory}, short-term~\cite{cowan2001magical}, and long-term categories~\cite{mandler1977long,vogt2007long,brady2008visual}. The visual memory game primarily captures long-term memory~\cite{isola2013makes}. Yet, given that HumanMem scores remain stable over various time delays~\cite{isola2013makes,lamem}, they are likely indicative of both short-term and long-term memory~\cite{borkin2015beyond,cowan2008differences}. Hence, our MachineMem measurer, which collects MachineMem scores, considers both short-term and long-term memory of machines. These aspects are captured by adjusting the training length in stage (b).

\paragraph{What images are more memorable to humans?}
Here, we briefly summarize the characteristics typically associated with human-memorable images:

$\bullet$ Images with large, iconic objects, usually in square or circular shapes and centered within the frame, tend to be more memorable. This suggests that a single iconic object makes an image more memorable than multiple objects.

$\bullet$ Images featuring human-related objects (such as persons, faces, body parts) and indoor scenes (like seats, floors, walls) have higher HumanMem scores, while outdoor scenes (such as trees, buildings, mountains, skies) generally contribute negatively.

$\bullet$ Bright, colorful images, especially those with contrasting colors or a predominantly red hue, are more memorable to humans.

$\bullet$ Simplicity in images often enhances memorability.

Conversely, images that deviate from these trends are generally less memorable. Furthermore, changes in other cognitive image properties (like aesthetics, interestingness, and emotional valence) show only weak correlations with HumanMem scores. As humans tend to construct simplified representations of the visual world for planning~\cite{ho2022people}, this may explain why simpler images are typically more memorable. For a more comprehensive understanding, we recommend readers to consult the relevant literature~\cite{isola2013makes,lamem,goetschalckx2019ganalyze,ho2022people}.

\paragraph{Memorization in DNNs.}
Previous research~\cite{feldman2020does,zhang2021understanding,arpit2017closer,feldman2020neural,toneva2018empirical,chatterjee2018learning} has explored the relationship between network memorization and aspects such as capacity, generalization, and robustness in supervised classification tasks. It is observed that more memorable samples are usually easy samples, while more forgettable samples tend to be hard. This is due to DNNs first learning patterns shared by common samples~\cite{kishida2019empirical,toneva2018empirical,feldman2020neural}. In a classification task, network memorization and data memorability are heavily influenced by labels and data distributions. In contrast, we focus on memorability, studying machine memorability of visual data generally, that is, without the constraints of labels and data distributions.

\begin{figure*}[!htb]
     \centering
     \includegraphics[width = \textwidth]
     {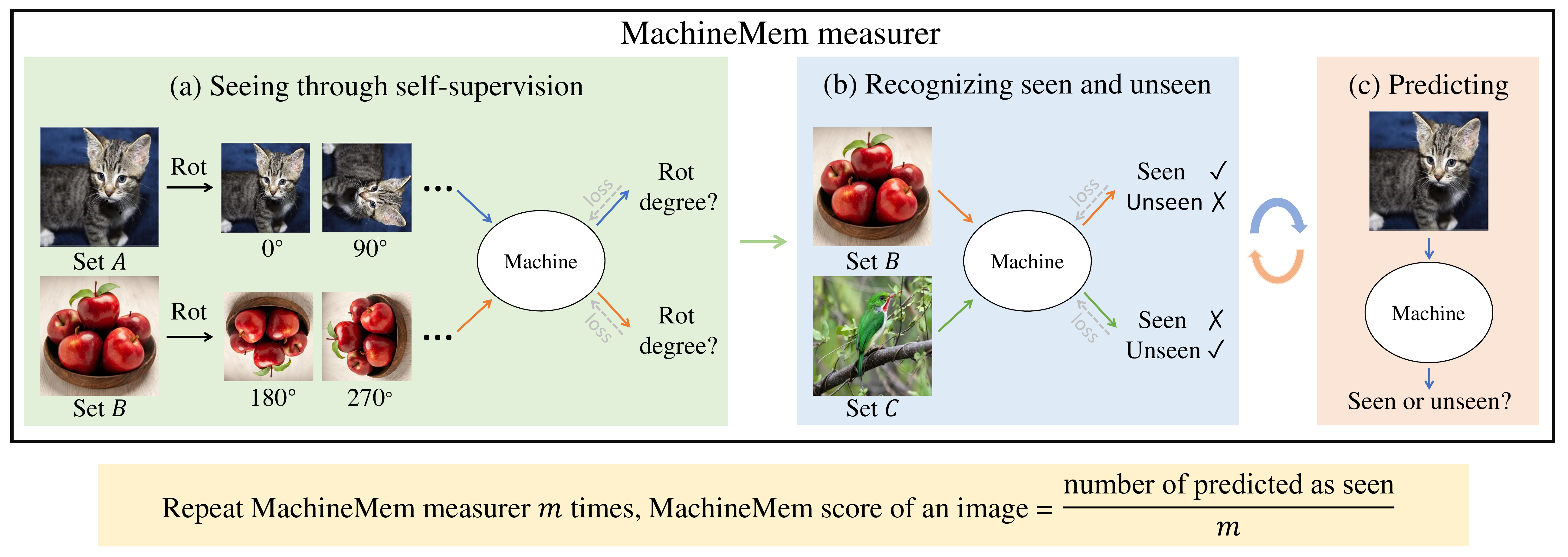}
     \caption{\textbf{A realization of the MachineMem measurer}. Our MachineMem measurer has 3 stages: (a) Seeing images through self-supervision, (b) Recognizing seen and unseen images, and (c) Predicting whether an image has been seen.Each image we present (cat, apple, and bird) symbolizes an image set ($A$, $B$, and $C$), each containing $n$ images. We focus on measuring MachineMem scores for set $A$ produced by an identical machine. In every episode of the MachineMem measurer, we randomly select sets $B$ and $C$ from an expansive dataset while keeping the cat set constant. The MachineMem scores of set $A$ are obtained by repeating the MachineMem measurer $m$ times. 
     }
     \label{fig:ppline}
\end{figure*}

\section{Measure Machine Memory}
\label{sec:machinemem}

We propose the MachineMem measurer as a pipeline to measure MachineMem scores of images. The design of the MachineMem measurer follows the key idea used in the visual memory game~\cite{isola2011makes}, that is, quantifying machine memory through a repeat detection task.  The visual memory game encapsulates three integral components: observe, repeat, and detect. Similarly, we structure the MachineMem measurer as a three-stage process, where each stage corresponds to one of these components. A conceptual diagram of the MachineMem measurer is presented in Figure~\ref{fig:ppline}.

Although humans are capable of observing and memorizing images without any feedback mechanism, machines, on the other hand, still lack this ability. Therefore, in stage (a), we adopt a self-supervision pretext task to guide machines to observe images. Following this, stage (b) instructs the machines to distinguish between observed and unobserved images, thereby equipping the machines to execute the repeat detection task in stage (c).

\paragraph{Seeing images through self-supervision.} 
In the endeavor to help machines observe and memorize images, supervision is indispensable. However, the supervision signal should be self-sufficient. In light of this, we contemplate employing a pretext task as supervision that satisfies three criteria: (1) it necessitates minimal structural modifications at the machine level, (2) it doesn't degrade or distort the input data, and (3) it allows machines to observe whole images rather than cropped segments. The rotation prediction self-supervision task~\cite{gidaris2018rotation} fulfills all these prerequisites and is therefore selected as the pretext task for stage (a).

Following common practice~\cite{gidaris2018rotation,deng2021does}, we define a set of rotation transformation functions $G= \{R_{r}(\bm{x})$\}, where $R_{r}$ is a rotation transformation function and $r$ are the rotation degrees $r\in\{0 \degree, 90 \degree, 180 \degree, 270 \degree \}$ . 
Rotation prediction is a multi-class (4 class here) classification task, where the goal is to predict which rotation degree has been applied to an input image $\bm{x}$.

The loss function is formulated as:
\begin{equation}
\mathcal{L}_{\textup{rot}} = \frac{1}{4} \left[ \sum_{  r \in \{ 0^{\circ}, 90^{\circ}, 180^{\circ}, 270^{\circ} \} } \mathcal{L}_{\textup{CE}}(\bm{y}_{r}, \bm{\theta}_{m}(R_{r}(\bm{x})))\right],
\end{equation}
where $\bm{y}_{r}$ is the one-hot label of $ r\in\{ 0^{\circ}, 90^{\circ}, 180^{\circ}, 270^{\circ} \}$. 
$\mathcal{L}_{\textup{CE}}$ denotes cross-entropy loss. 
$\bm{m}$ is a machine parameterized by $\bm{\theta}_{m}$.

Stage (a) uses two sets (sets $A$ and $B$) of images, where images in both two sets  are labeled as seen. Half of them (set $B$) go to stage (b) and the other half (set $A$) go to stage (c), as shown in Figure~\ref{fig:ppline}. We force machines to achieve good performance (top-1 accuracy $\geq$ 80 \%) on the rotation prediction task. By default, machines without pre-training are trained for 60 epochs in this stage.

\paragraph{Recognizing seen and unseen images.} 
This stage aims to teach machines to recognize seen and unseen images. We use a set of seen images (set $B$) that has been used in stage (a) and sample a set of unseen images (set $C$) from a large-scale dataset.
A binary classification task targeted at recognizing seen and unseen images is employed. We replace the 4-way classification layer with a new 2-way linear classification head. The loss function is expressed as:
\begin{equation}
\mathcal{L}_{\textup{seen}} = \frac{1}{2} \left[ \sum_{  r \in \{ \rm{seen, unseen} \} } \mathcal{L}_{\textup{CE}}(\bm{y}_{l}, \bm{\theta}_{m}(\bm{x}))\right],
\end{equation}
where $\bm{y}_{l}$ denotes the one-hot label of $r\in\{\rm{seen, unseen}\}$, CE is cross-entropy loss, and $\bm{m}$ stands for a machine parameterized by $\bm{\theta}_{m}$.

In our default setting, stage (b) lasts for 10 epochs, and the machine will enter stage (c) upon finishing each epoch.

\paragraph{Predicting whether an image has been seen.}
During the final stage, we utilize a set of previously 'seen' images (set $A$) that were not involved in stage (b) for measurement purposes. Here, the machine's task is to discern whether a given image has been seen before, thereby replicating the repeat detection tasks we use to evaluate human memory capabilities.



Given the inherent uncertainty in learning-based systems~\cite{hullermeier2021aleatoric}, it is imperative to ensure that the results generated during this stage are both meaningful and reliable. To this end, we employ the calibration error metric~\cite{guo2017calibration} as an assessment tool to gauge the reliability of the machine's predictions. Lower calibration errors are indicative of models with a higher degree of reliability and result accuracy. Specifically, we employ the RMS calibration error~\cite{hendrycks2018deep} and adaptive binning~\cite{nguyen2015posterior} techniques to measure this.


To improve the robustness of our method, we have further enhanced our original approach to incorporate a held-out set of images for the assessment of calibration quality in both seen and unseen image categories. We have validated this enhanced strategy across several neural network models and found that the MachineMem scores produced align closely with our original results, which were based solely on seen images. This was supported by a strong Spearman's correlation ($\rho > 0.6$) across all tested machines. These findings substantiate that assessing calibration quality using seen images alone is adequate for reliable measurements.

The interchange between stage (b) and stage (c) is iterated, allowing us to gather multiple measurements from the images within set $A$. The final result from a single MachineMem measurer episode is chosen based on the iteration that yields the lowest calibration error. This approach also captures both the short-term and long-term memory capabilities of the machines.


\paragraph{Obtaining MachineMem scores.}
Drawing from the HumanMem scores approach~\cite{isola2011makes}, we define the MachineMem score of an image as the ratio of the number of seen predictions to the total number of MachineMem episodes.

During each MachineMem measurer episode, besides set $A$ which is destined for stage (c) for score computation, we randomly select two other sets of images (set $B$ and set $C$) from a dataset, each containing $n$ images. This randomized selection process is designed to ensure that the MachineMem measurer accurately reflects the machine's memory capabilities rather than fitting specific distributions. We default $n$ to 500. To calculate the MachineMem scores for set $A$, we repeat the MachineMem process $m$ times, where $m$ is set to 100, mirroring the average number of participants involved in HumanMem score collection. The MachineMem scores for set $A$ are thus obtained after repeating the MachineMem measurer $m$ times.

We have collected and labeled MachineMem scores for all images in the LaMem dataset (totaling 58,741 images). The average MachineMem score is 0.680 (SD 0.070, min 0.39, max 0.91), which contrasts with the average HumanMem score of 0.756 (SD 0.123, min 0.20, max 1.0). By default, we employ a ResNet-50 as our basic machine and use the scores produced by this machine to represent the MachineMem scores. We have also endeavored to evaluate the memory characteristics of various other machine models, with their respective training specifics and analyses discussed in the subsequent sections.



\paragraph{Types of machines and experimental details.}
In our research, we investigate the memory characteristics of 11 distinct machines. These machines are categorized into four groups, namely conventional machines (comprising linear classifier and SVM~\cite{cowan2001magical}), classic CNNs (such as AlexNet~\cite{alexnet} and VGG~\cite{vgg}), modern CNNs (including ResNet-18, ResNet-50~\cite{resnet}, ResNet-152, WRN-50-2~\cite{Zagoruyko2016WRN}, and DenseNet121~\cite{huang2019convolutional}), and recent ViTs (like XCiT-T~\cite{ali2021xcit} and MaxViT-T~\cite{tu2022maxvit}).

Except for the number of training epochs in stage (a) and the corresponding learning rate, all training parameters remain consistent across all machine types and pre-training methods. We have made these adjustments to ensure machines are able to achieve satisfactory performance levels (top-1 accuracy $\geq$ 80\%) during stage (a).

Our MachineMem measurer is trained solely on a single GPU, with a batch size of 1 to parallel the visual repeat game settings. We employ SGD as our optimization algorithm and use a cosine learning schedule for our training process. The settings for weight decay and momentum are 0.0001 and 0.9, respectively. The specifics for the training epochs in stage (a) and learning rates for all machine models are as follows:

Conventional machines: Training epochs for stage (a) are set at 60, with a learning rate of 0.01.

Classic CNNs: Training epochs for stage (a) are set at 70, with a learning rate of 0.0005.

Modern CNNs: Training epochs for stage (a) are set at 60, with a learning rate of 0.01.

ViTs: Training epochs for stage (a) are set at 70, with a learning rate of 0.0005.

ResNet-50 with pre-training: Training epochs for stage (a) are set at 30, with a learning rate of 0.01.

\section{Predict MachineMem Scores}
Collecting MachineMem scores with the MachineMem measurer can be a time-consuming process, often requiring several hours to generate scores for 500 images. In response to this challenge, we aspire to train a robust regression model capable of predicting MachineMem scores in real-time. This task aligns with the prediction of HumanMem scores, prompting us to revisit and enhance approaches tailored towards predicting HumanMem scores.

Past research~\cite{fajtl2018amnet,perera2019image} has demonstrated that a modified ResNet-50 regression model (with adjustments to the final layer to accommodate regression tasks) can deliver satisfactory performance in predicting HumanMem scores. This model is trained utilizing dropout~\cite{srivastava2014dropout} and RandomResizedCrop~\cite{szegedy2015going}. With this particular training setup, complemented by an ImageNet-supervised pre-training initialization, this straightforward ResNet-50 regression model can attain a Spearman's correlation, $\rho$, of 0.63 when predicting HumanMem scores. For comparison, the human consistency~\cite{lamem} registers at $\rho = 0.68$, while the state-of-the-art result~\cite{perera2019image} reaches $\rho = 0.67$.

Our findings suggest that a superior correlation can be accomplished in predicting HumanMem scores by employing self-supervised pre-training and strong data augmentations. Specifically, we transfer the knowledge from the proficiently trained MoCo v2~\cite{chen2020improved} as opposed to supervised ImageNet classification. At the data level, we substitute RandomResizedCrop~\cite{szegedy2015going} with CropMix~\cite{han2022cropmix}, while integrating Random erasing~\cite{zhong2020random} and Horizontal flip applied in a YOCO manner~\cite{han2022yoco}. This results in a ResNet-50 regressor that attains $\rho = 0.69$ in predicting HumanMem scores, surpassing even human consistency! We refer to this model as the enhanced ResNet-50 regression model. 

In the subsequent step, we aspire to train a regression model capable of predicting MachineMem scores that align as closely as possible with those derived from empirical observations of 100 trials from the MachineMem measurer. We employ and train this enhanced ResNet-50 regression model for the task of predicting MachineMem scores. We randomly select 1000 images as the test set, using all remaining images as the training set. This model also achieves a $\rho = 0.69$ in predicting MachineMem scores. We designate our model as the MachineMem predictor.

\begin{figure*}[!htb]
     \centering
    \begin{minipage}[t]{0.24\linewidth} 
    \centering 
\includegraphics[width=4.6cm]{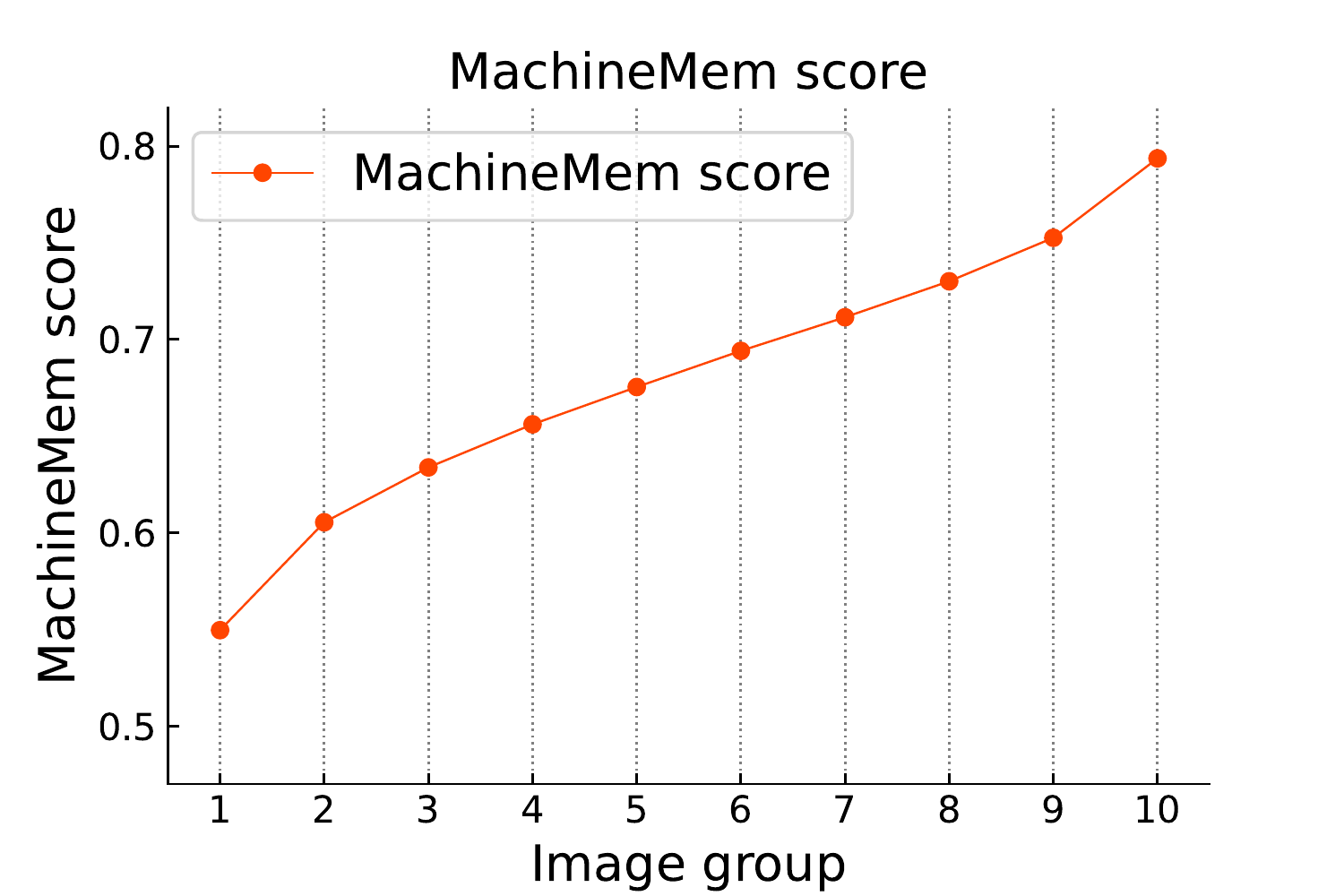}
  \end{minipage}   
    \begin{minipage}[t]{0.24\linewidth} 
    \centering 
\includegraphics[width=4.6cm]{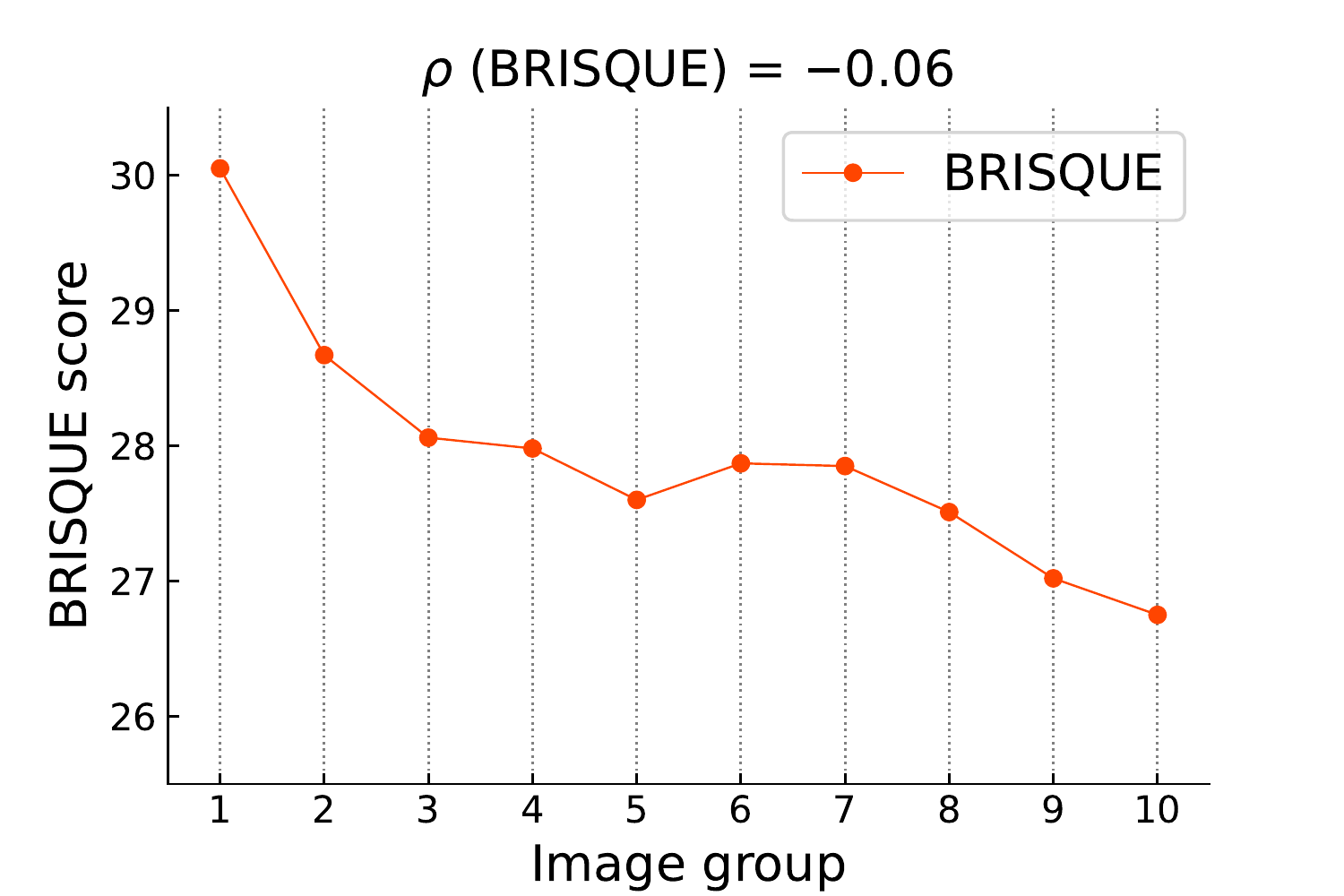}
  \end{minipage}
    \begin{minipage}[t]{0.24\linewidth} 
    \centering 
\includegraphics[width=4.6cm]{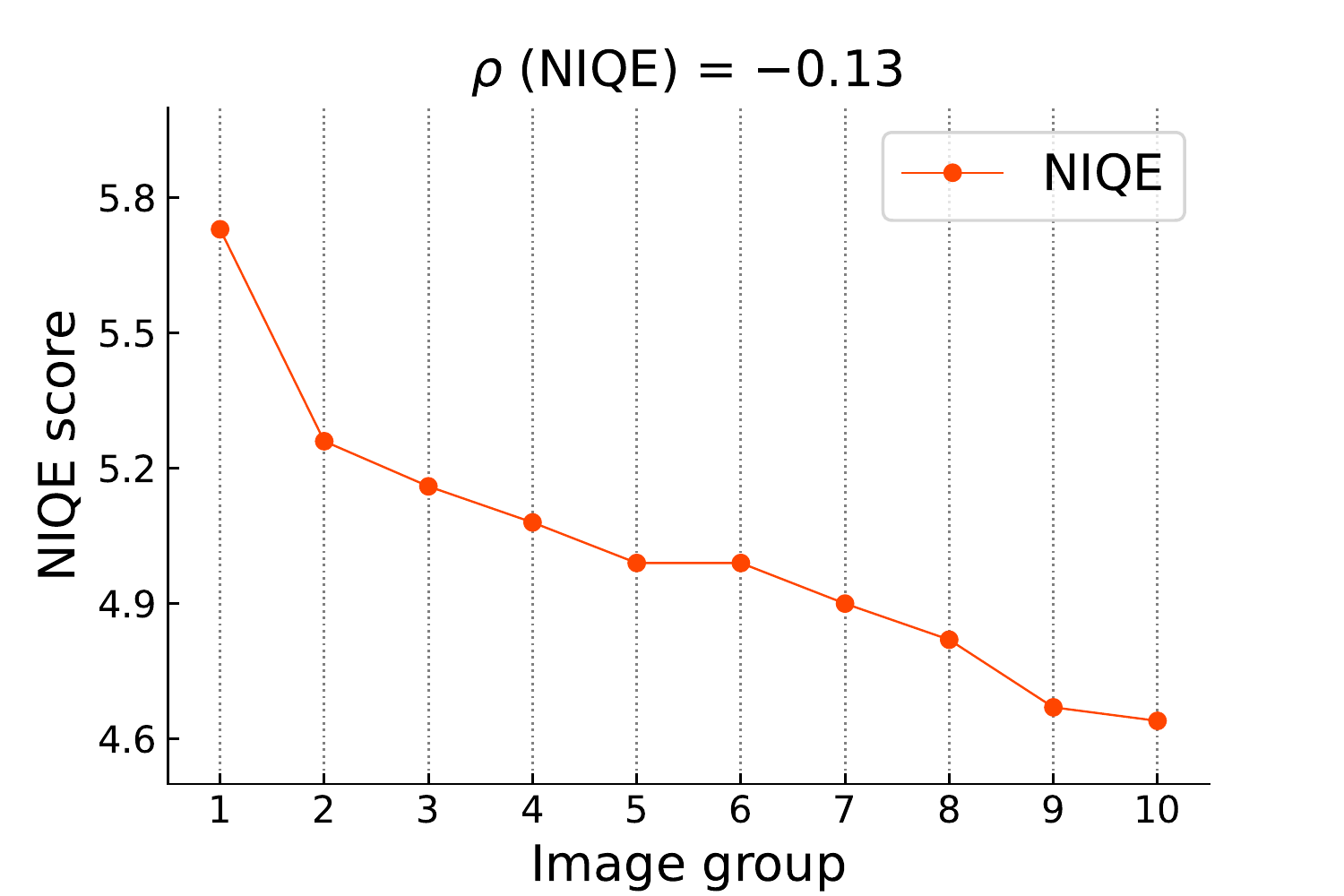}
  \end{minipage}   
    \begin{minipage}[t]{0.24\linewidth} 
    \centering 
\includegraphics[width=4.6cm]{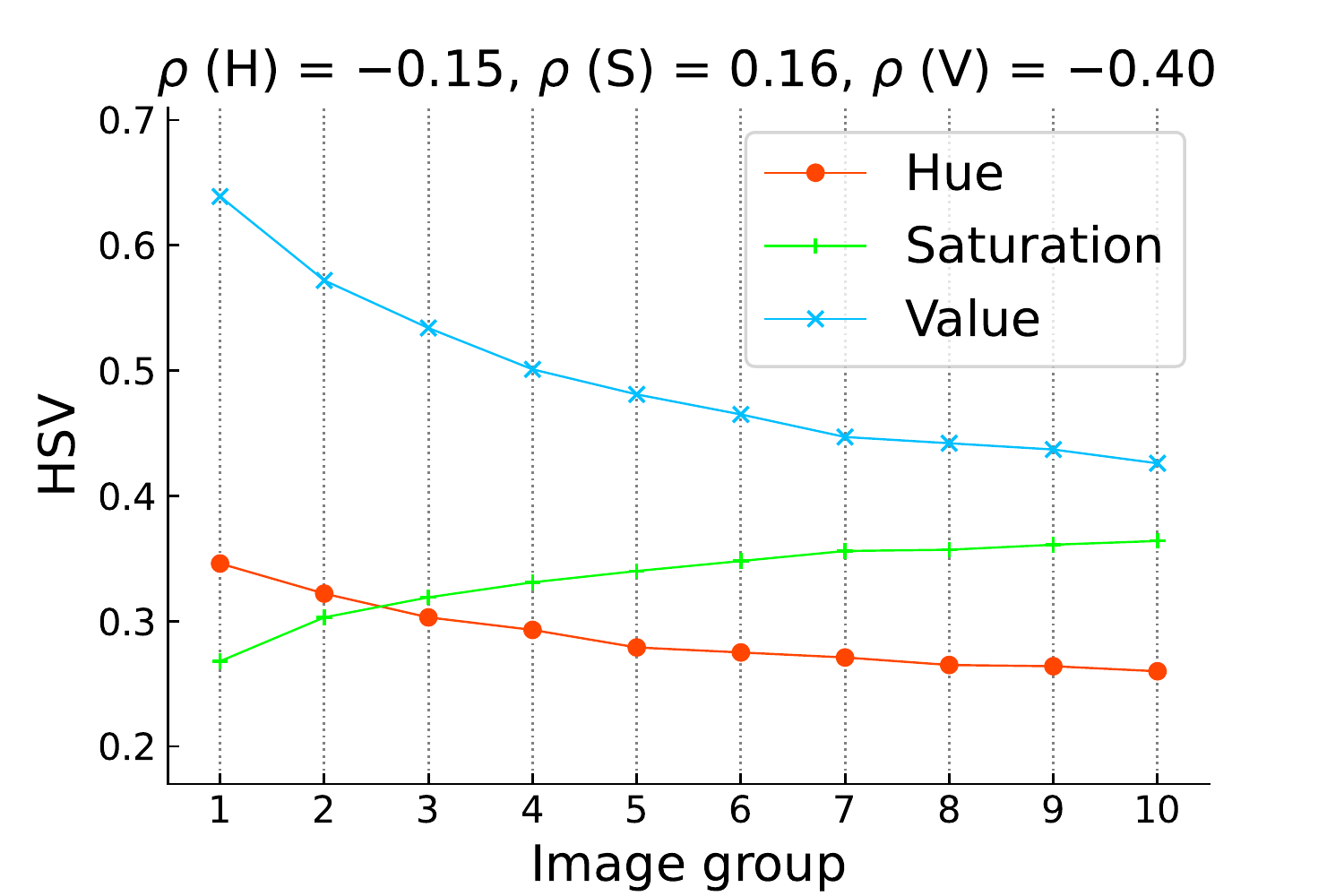}
  \end{minipage}  
    \begin{minipage}[t]{0.24\linewidth} 
    \centering 
\includegraphics[width=4.6cm]{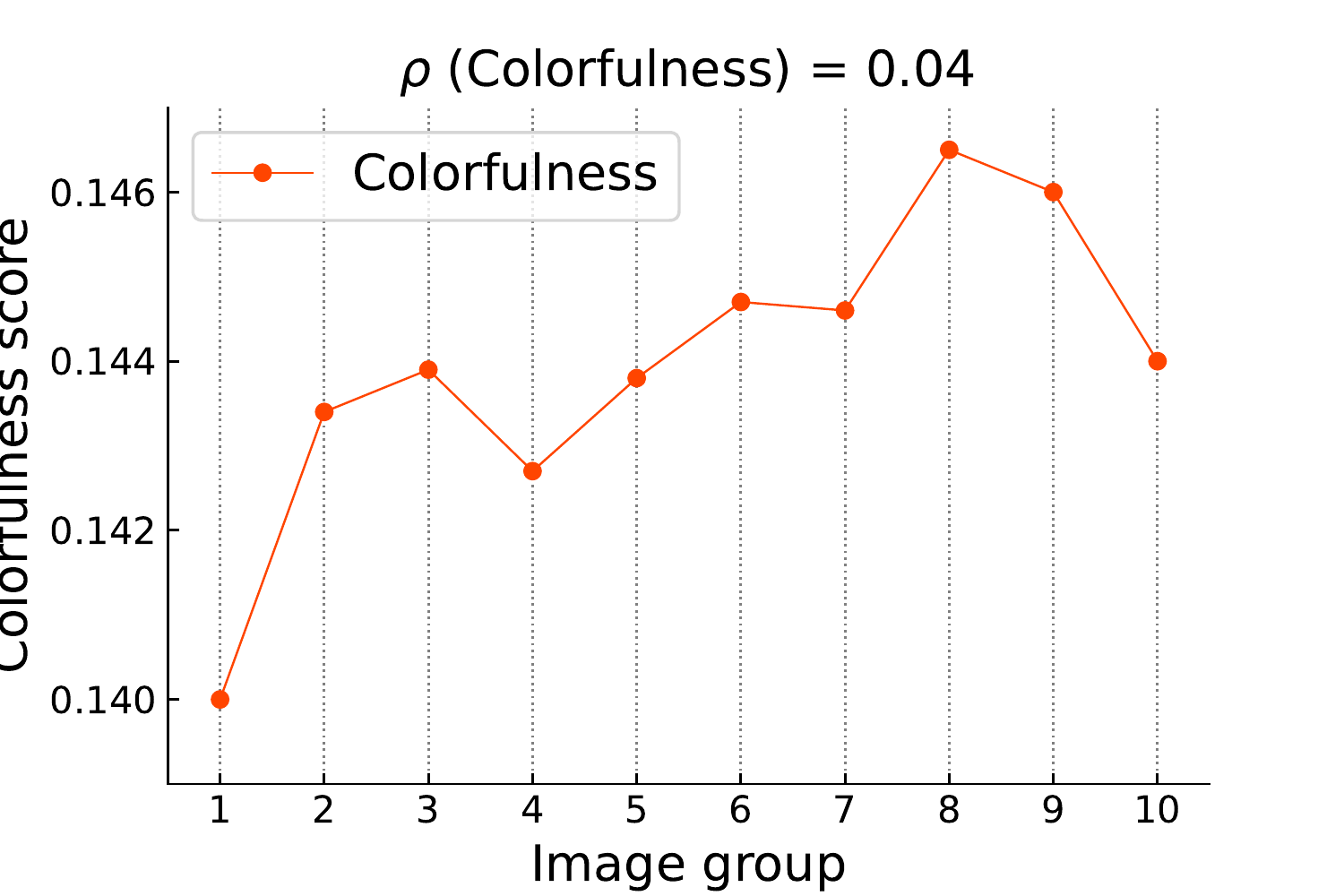}
  \end{minipage}   
    \begin{minipage}[t]{0.24\linewidth} 
    \centering 
\includegraphics[width=4.6cm]{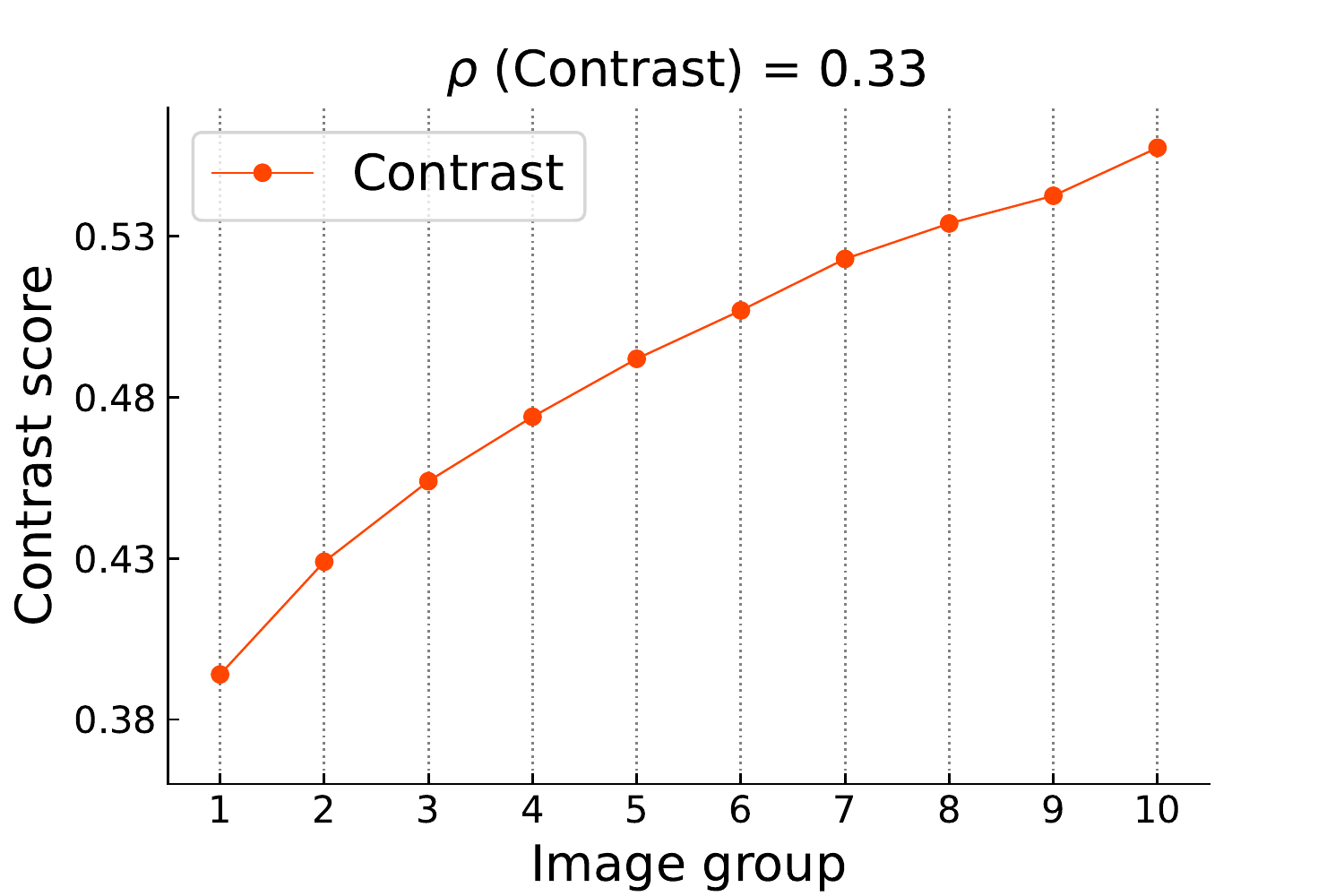}
  \end{minipage}
    \begin{minipage}[t]{0.24\linewidth} 
    \centering 
\includegraphics[width=4.6cm]{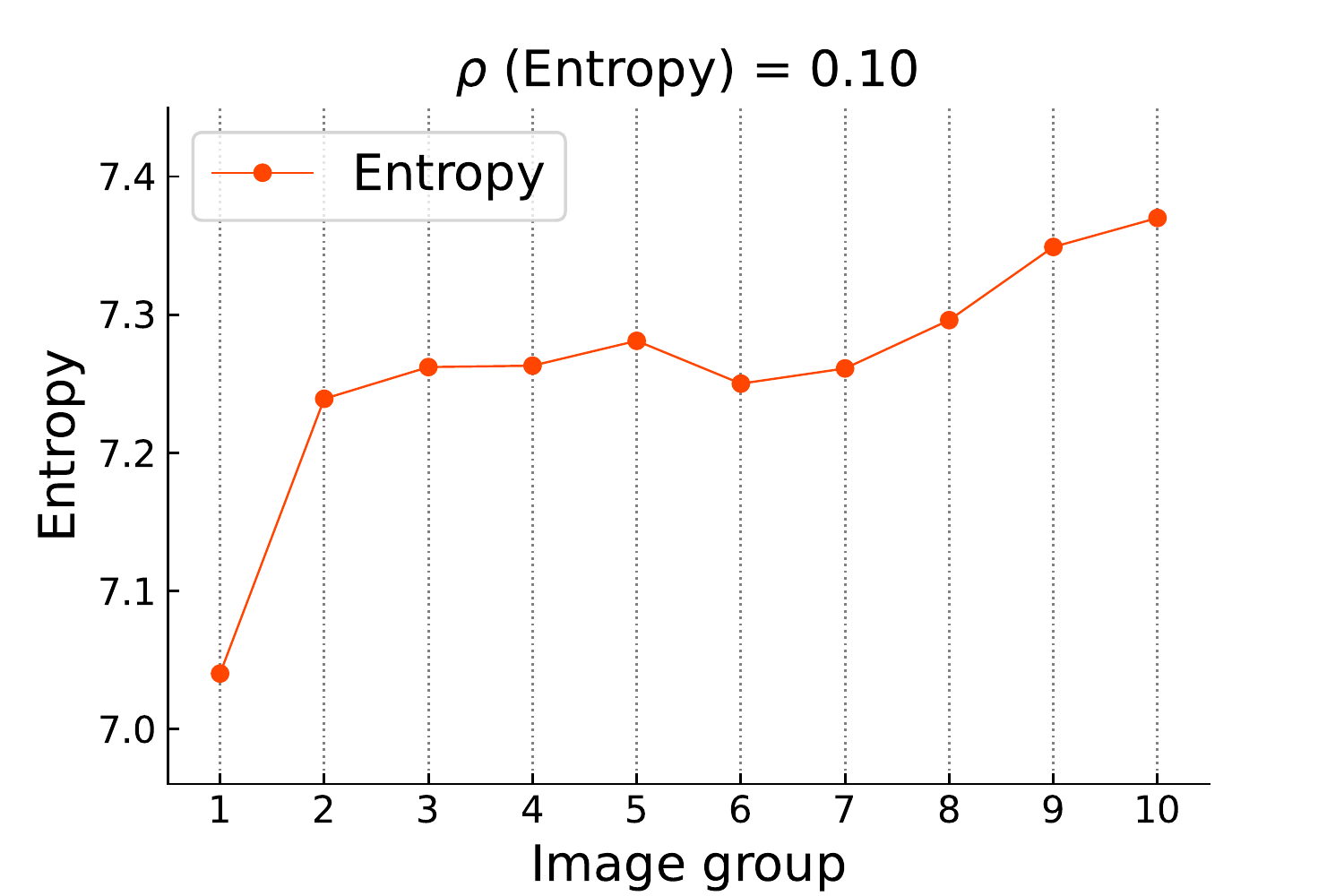}
  \end{minipage}   
    \begin{minipage}[t]{0.24\linewidth} 
    \centering 
\includegraphics[width=4.6cm]{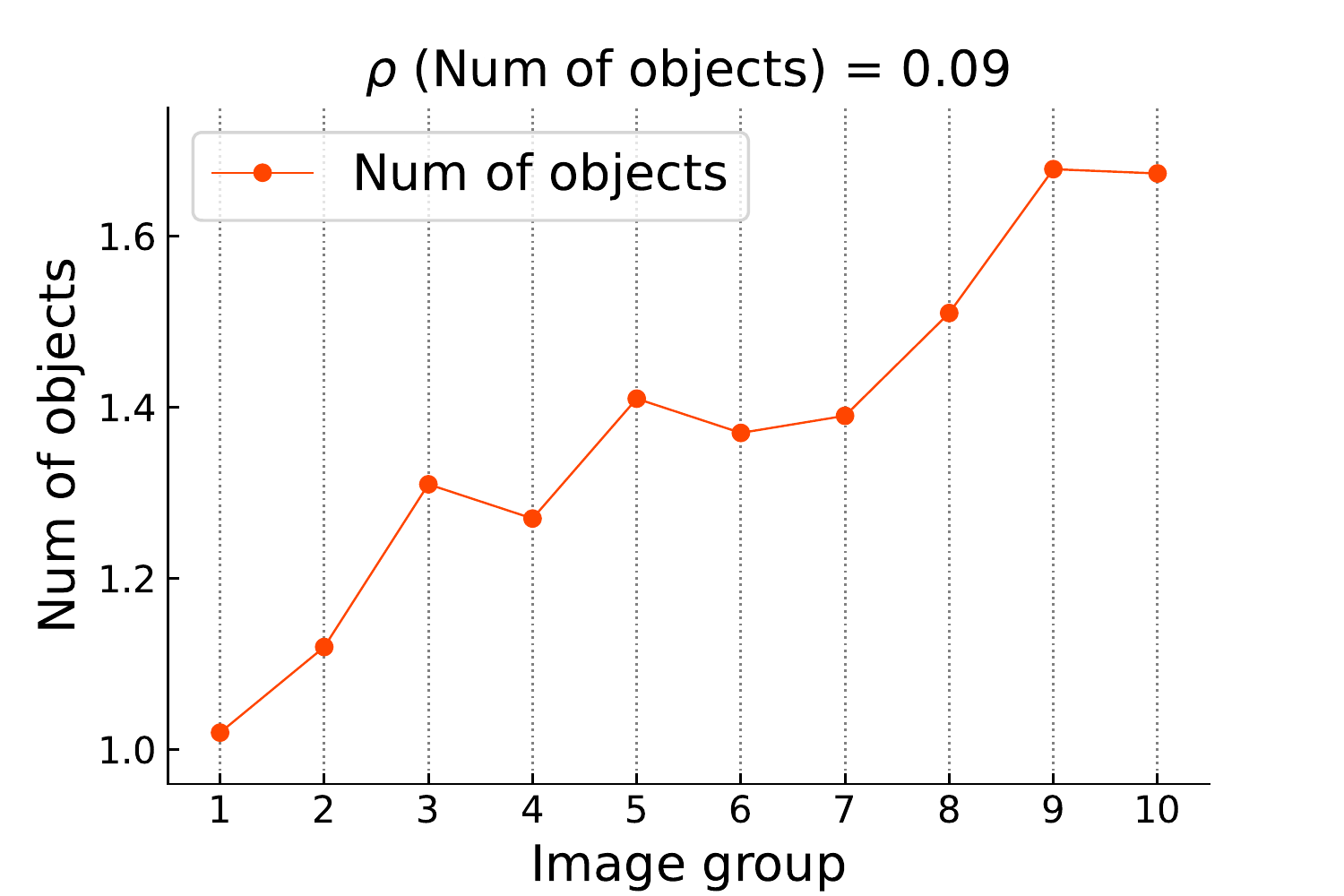}
  \end{minipage}    
    \begin{minipage}[t]{0.24\linewidth} 
    \centering 
\includegraphics[width=4.6cm]{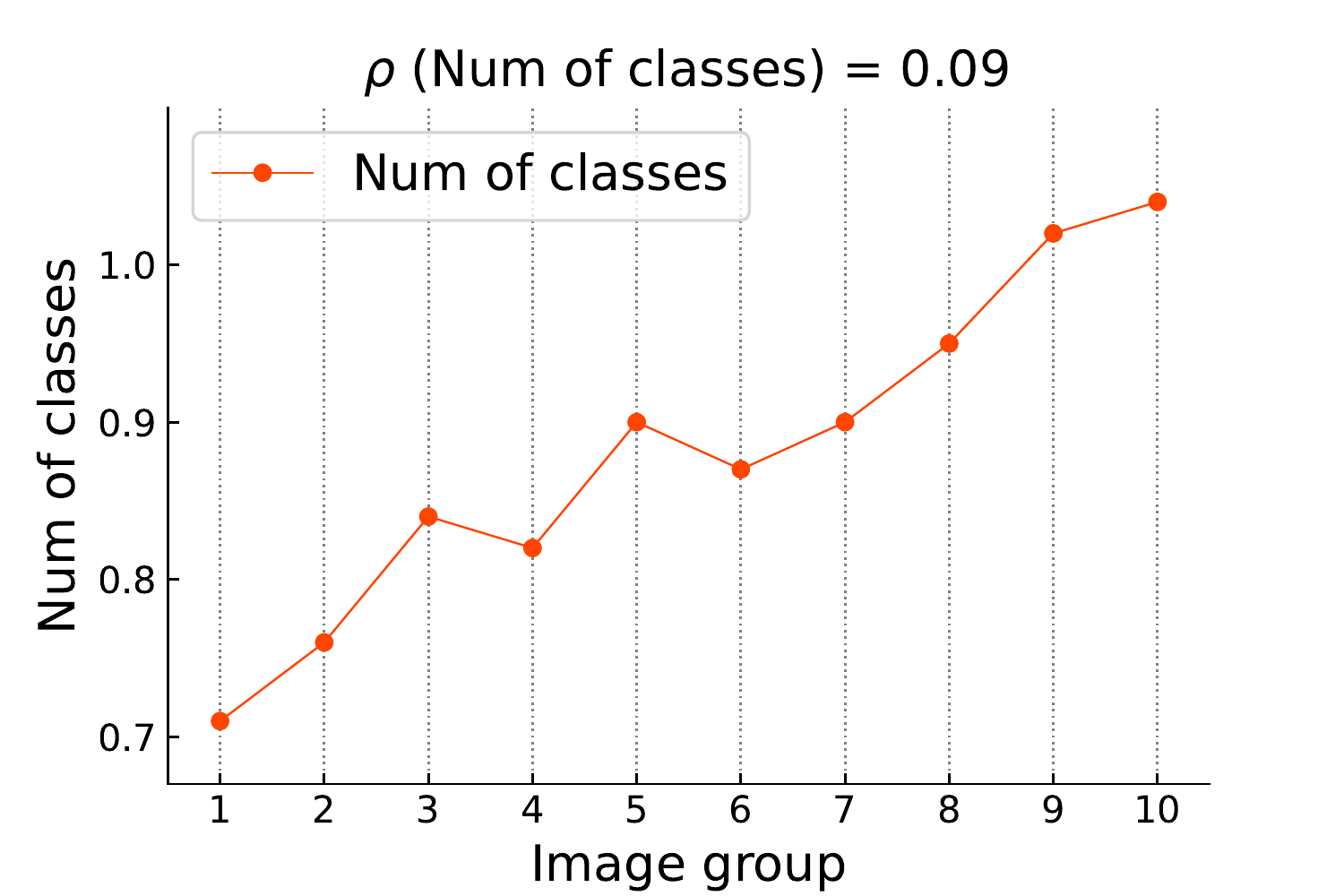}
  \end{minipage}    
      \begin{minipage}[t]{0.24\linewidth} 
    \centering 
\includegraphics[width=4.6cm]{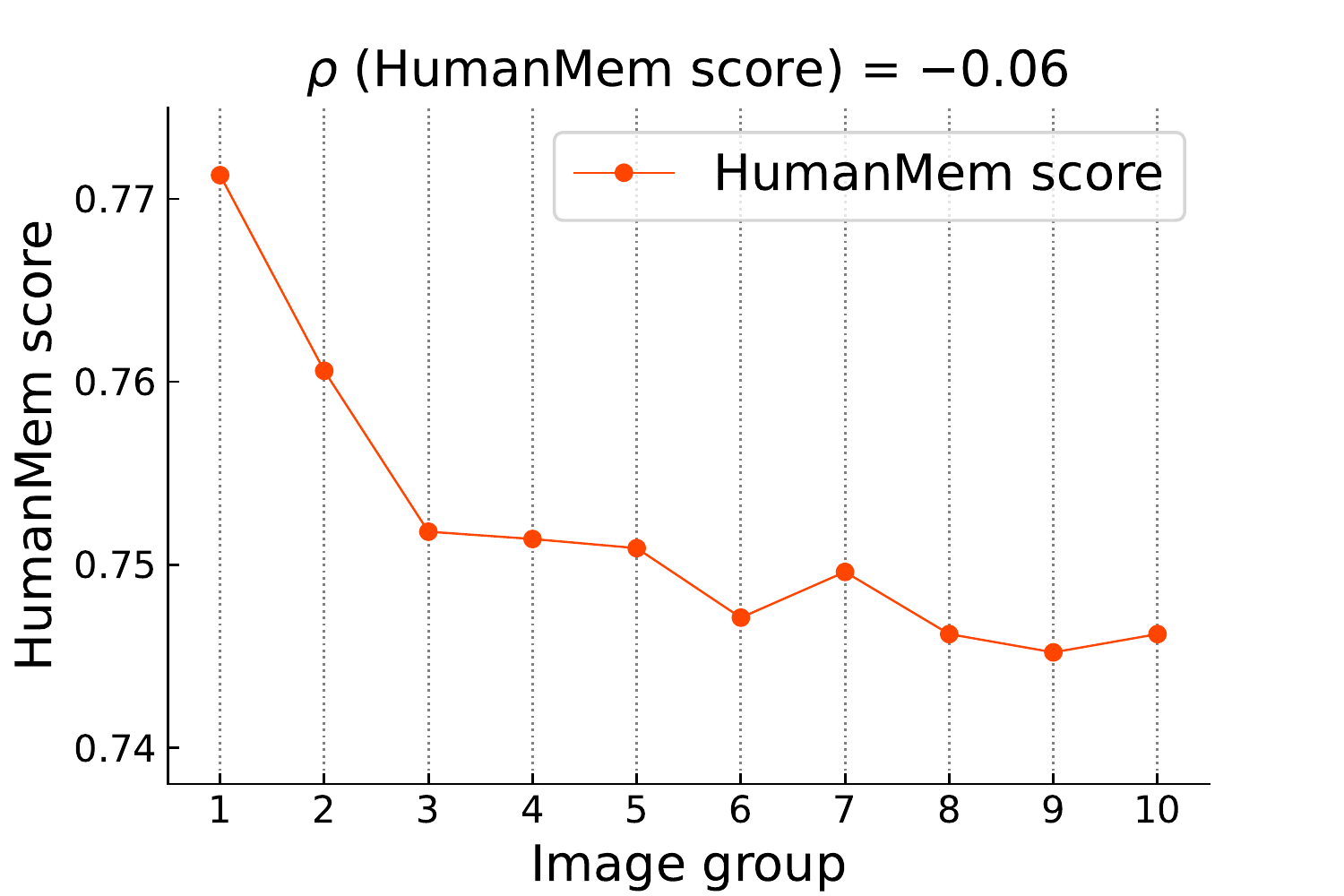}
  \end{minipage}    
      \begin{minipage}[t]{0.24\linewidth} 
    \centering 
\includegraphics[width=4.6cm]{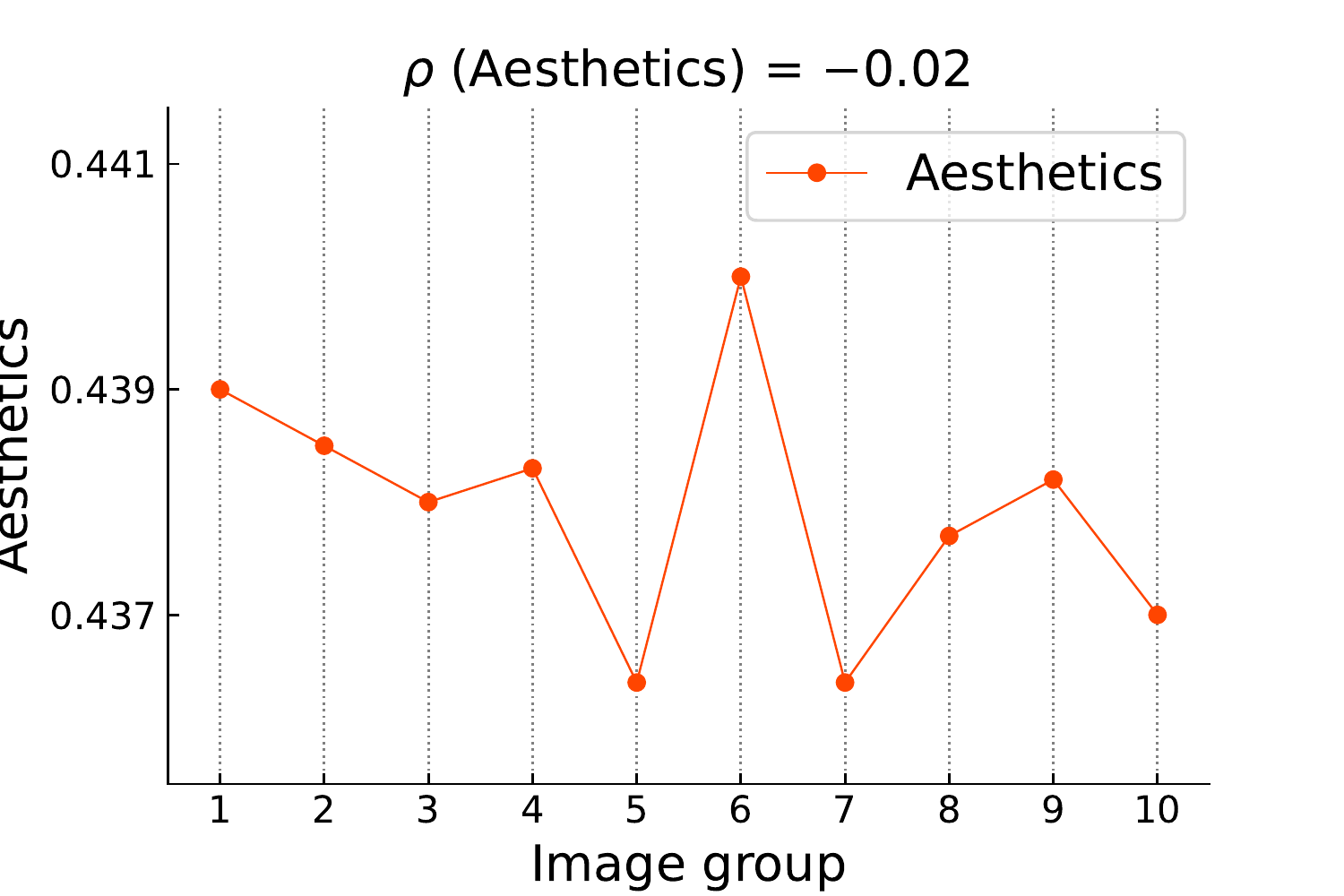}
  \end{minipage}    
      \begin{minipage}[t]{0.24\linewidth} 
    \centering 
\includegraphics[width=4.6cm]{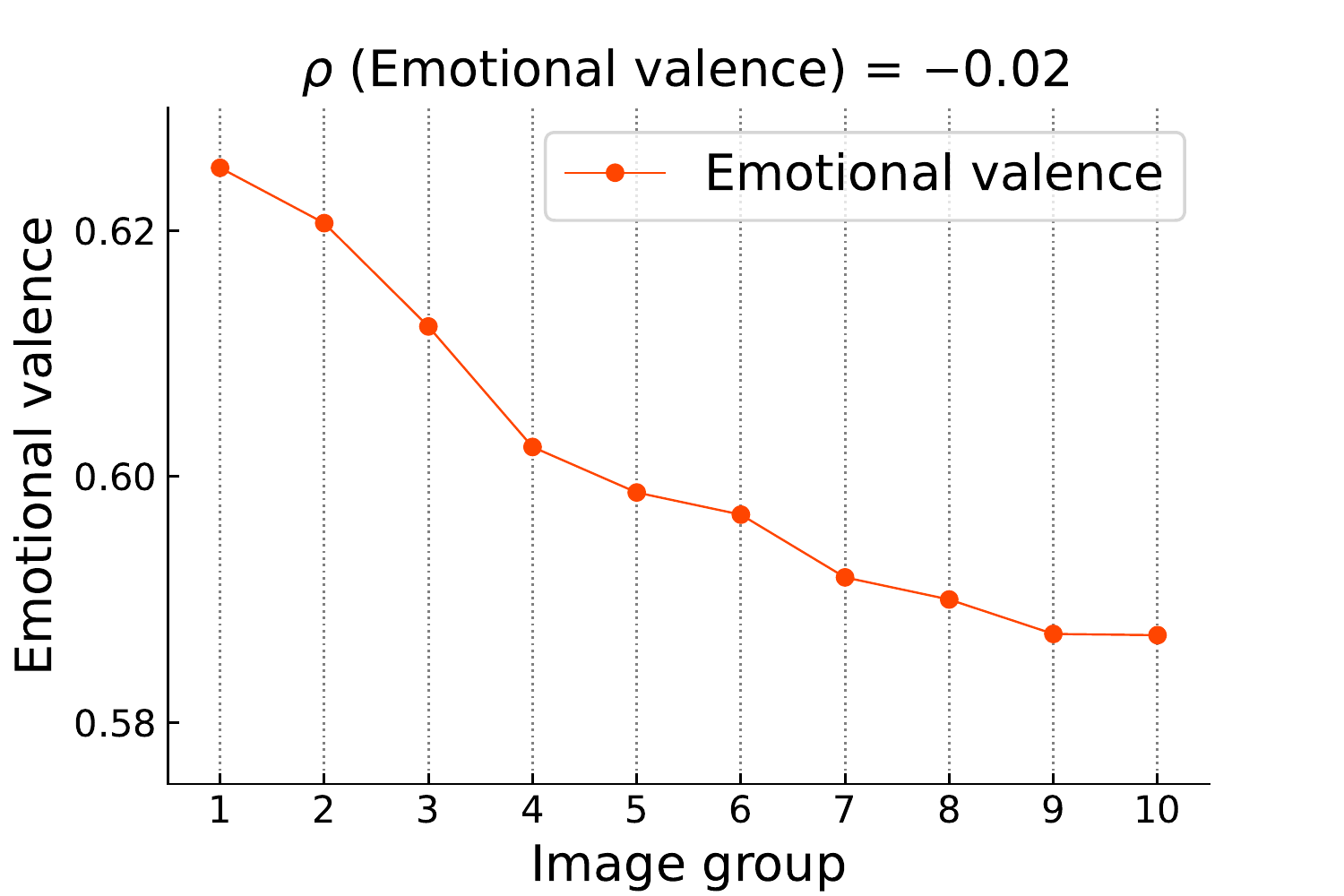}
  \end{minipage}    
    \caption{\textbf{Relation between image groups and varying image attributes}. We find value and contrast to be the two most significant attributes that correlate moderately ($\rho \geq 0.3$) with MachineMem scores. On the other hand, attributes such as hue and saturation show a weak correlation ($ 0.15 \leq \rho < 0.3$) with MachineMem scores. Other factors, such as NIQE, entropy, number of objects, and number of classes, demonstrate a very weak correlation ($ 0.08 \leq \rho < 0.15$) with MachineMem scores. These findings are based on Spearman's correlation ($\rho$) computed from the entire data set.
    }
    \label{fig:lamem}
\end{figure*}

\section{What Makes an Image Memorable to Machines? }
This section aims to analyze MachineMem scores in order to understand what factors contribute to an image's memorability for machines. We present some sample images in Figure~\ref{fig:lowtohigh}, first analyzing the relationship between MachineMem scores and 13 image attributes. With the aid of the MachineMem predictor, we predict MachineMem scores of all ImageNet~\cite{imagenet} and  Places~\cite{zhou2017places} training images to analyze which classes and scenes are most and least memorable to machines. Additionally, we employ the GANalyze~\cite{goetschalckx2019ganalyze}, capable of adjusting an image to generate more or less memorable versions, as a means to discover hidden trends that could potentially influence MachineMem scores. In conjunction with GANalyze, we conduct a comparative study between machine memorability and human memorability, and further explore what images are more memorable to other machines.

\subsection{Quantitative analysis}
Do image attributes adequately determine MachineMem scores? Here we examine 13 image attributes, roughly grouped into 4 categories, each focusing on different measurements. Based on MachineMem scores of images, we sort all LaMem images and organize them into 10 groups, from the group with the lowest mean MachineMem scores to the highest. Each group contains approximately 5870 images. Spearman's correlation results are computed based on individual data (all 58741 images). Figure~\ref{fig:lamem} presents plots illustrating the relationship between image groups and varying image attributes.

\textbf{Image quality.} We employ two metrics, NIQE~\cite{mittal2012making} and BRISQUE~\cite{mittal2012no}, to assess image quality. Lower NIQE and BRISQUE values suggest better perceptual quality, indicating fewer distortions.

Both BRISQUE and NIQE demonstrate a very weak correlation with MachineMem scores ($\rho$ = $-0.06$ and $-0.13$, respectively). NIQE shows a relatively stronger correlation, possibly because BRISQUE, which involves human subjective measurements, correlates more strongly with human perception.

\textbf{Pixel Statistics.} We investigate the correlation between MachineMem scores and basic pixel statistics. Hue, saturation, and value from the HSV color space~\cite{max2005computer} are measured, along with colorfulness~\cite{hasler2003measuring}, contrast~\cite{matkovic2005global}, and entropy.

Interestingly, value and contrast show substantial correlations with MachineMem scores ($\rho$ = $-0.40$ and $-0.33$, respectively). Deep color and strong contrast are two significant factors that make an image memorable to machines. Hue and saturation are weakly correlated with MachineMem scores ($\rho$ = $-0.15$ and $0.16$, respectively). Entropy exhibits a very weak correlation with MachineMem scores ($\rho$ = $0.10$). However, as presented in Figure~\ref{fig:lamem}, the group with the lowest MachineMem scores, \ie{}, group 1, displays very low entropy. Images with very low MachineMem scores often lack contrast or have a light color background (see Figure~\ref{fig:lowtohigh}), and therefore tend to have low entropy. Furthermore, colorfulness seems to have no clear correlation with MachineMem scores ($\rho$ = $0.04$), except for the fact that group 1 scores very low in terms of colorfulness.

\textbf{Object Statistics.} We measure the number of objects and the number of classes (unique objects) within an image. A pre-trained YOLOv4~\cite{bochkovskiy2020yolov4} is employed as the object detector. 

Both number of objects and classes are very weakly correlated with MachineMem scores (same $\rho$ = $0.09$). By excluding data with 0 objects/classes, their correlations with MachineMem scores are still very weak (same $\rho$ = $0.10$).

\textbf{Cognitive Image Property.} Our analysis considers three cognitive image properties: HumanMem scores, aesthetics, and emotional valence. The HumanMem scores are obtained from the LaMem dataset. To measure aesthetics, we utilize a pre-trained LIMA model~\cite{talebi2018nima}. We use the emotional valence predictor from the GANalyze tool~\cite{goetschalckx2019ganalyze} for determining emotional valence.

The correlation between HumanMem scores and MachineMem scores is very weak ($\rho$ = $-0.06$). Similarly, other cognitive image properties, such as aesthetics and emotional valence, exhibit negligible correlation with MachineMem scores (both with $\rho$ = $-0.02$). These findings suggest that MachineMem scores represent a unique image property that is distinct from other image properties.

In conclusion, unlike human memory, which is largely driven by semantics, machines, devoid of such common sense, tend to emphasize more on basic pixel statistics.

\subsection{What classes are more or less memorable?} 
\label{subsec:classes}
\begin{figure*}[!htb]
     \centering
     \includegraphics[width = \textwidth]
     {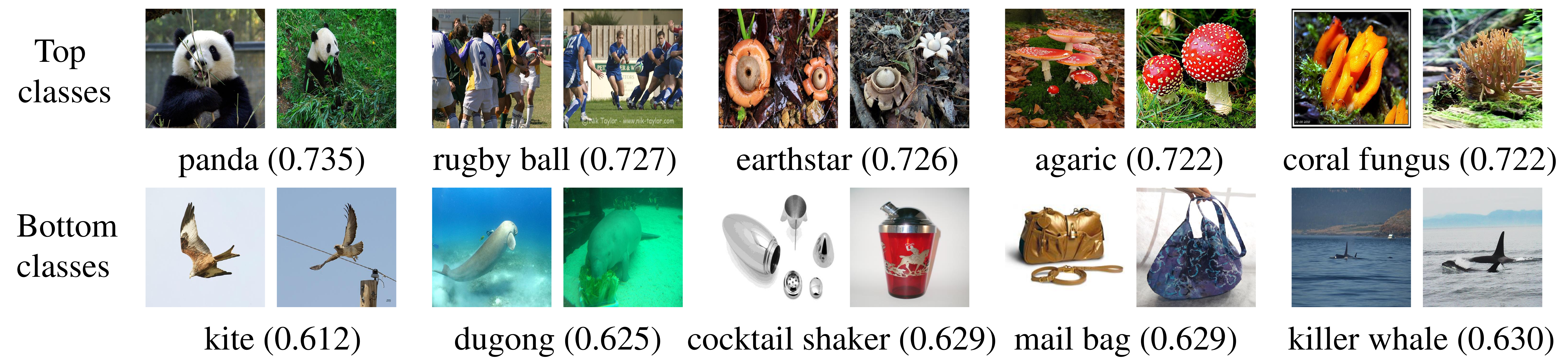}
     \caption{\textbf{ImageNet classes sorted by their mean MachineMem scores}. We report the top-5 and bot-5 classes and their mean MachineMem scores. It appears that the classes ranking highest commonly exhibit lower values paired with pronounced contrast. To illustrate, images of pandas consistently feature a mix of both white and black hues, often juxtaposed against a green background, thus enhancing the overall contrast. Conversely, classes ranked at the bottom predominantly showcase lighter backgrounds occupying a substantial proportion of the pixel space. 
     }
     \label{fig:imagenet}
\end{figure*}

Do images belonging to certain classes tend to be more or less memorable to machines? We use the MachineMem predictor to predict  MachineMem scores of all ImageNet training images to obtain mean MachineMem scores of 1000 (ImageNet) classes. Figure~\ref{fig:imagenet} summarizes the top and bottom classes. By analyzing gained results, we find the answer to be yes: Classes potentially containing light backgrounds are usually less memorable to machines, for instance, classes related to sea or sky. Classes that have strong contrast, high value, and multiple objects tend to have high MachineMem scores. 

\subsection{What scenes are more or less memorable?}
\begin{figure*}[!htb]
     \centering
     \includegraphics[width = \textwidth]
     {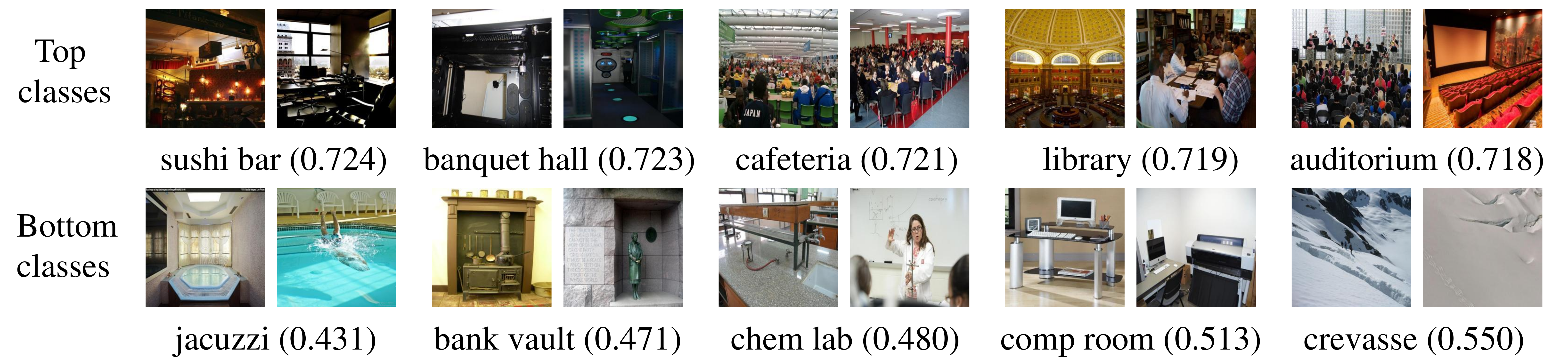}
     \caption{\textbf{Places scenes sorted by their mean MachineMem scores}. The top-5 and bot-5 scenes and their mean MahcineMem scores are reported. As observed in the ImageNet classes, the scenes with higher MahcineMem scores generally exhibit lower value, higher contrast, and contain multiple objects. Alternatively, the scene with lower MahcineMem scores often feature white walls and while outdoor scenes.
     }
     \label{fig:place}
\end{figure*}
Do the trends observed from ImageNet classes apply to scenes as well? To answer this, we present the top and bottom Places scenes in Figure~\ref{fig:place}. We utilized the MachineMem predictor to estimate MachineMem scores for all the training images in Places365~\cite{zhou2017places}, thus enabling us to compute average MachineMem scores for 365 scenes. The results indicate that the patterns identified in classes/objects are also evident in scenes, suggesting that this trend is broadly applicable to visual data.

\subsection{Can more or less memorable classes be semantically grouped according to a hierarchical structure?}

We utilized ImageNet's supercategories to delve into this question. Table~\ref{tab:supercategory} outlines the top-five and bottom-five ImageNet supercategories, together with their average MachineMem scores. These findings align with our class-level observations, confirming that memorable classes can indeed be semantically grouped according to a hierarchical structure.

\begin{table}[!htb]
  \centering
  \fontsize{7}{3}\selectfont
    \begin{tabular}{lc|c|c|c|c}
    \toprule
     \Rows{Top-5}&  basidiomycete  & procyonid &  player   & marketplace&  fungus\cr
    & 0.722 &  0.720 &  0.716 &  0.716 &  0.714 \cr
    \midrule
     \Rows{Bot-5}&  rescue equipment & computer & reservoir   & sailing vessel &  hawk\cr
    & 0.606 &  0.607 &  0.608 &  0.612 &  0.612 \cr    
    \bottomrule
    \end{tabular}
    \caption{ImageNet supercategories sorted by their mean MachineMem scores. We report top-5 and bot-5 supercategory and their mean MahcineMem scores.}
    \label{tab:supercategory}
\end{table}

\begin{figure*}[!htb]
     \centering
     \includegraphics[width = 16cm]
     {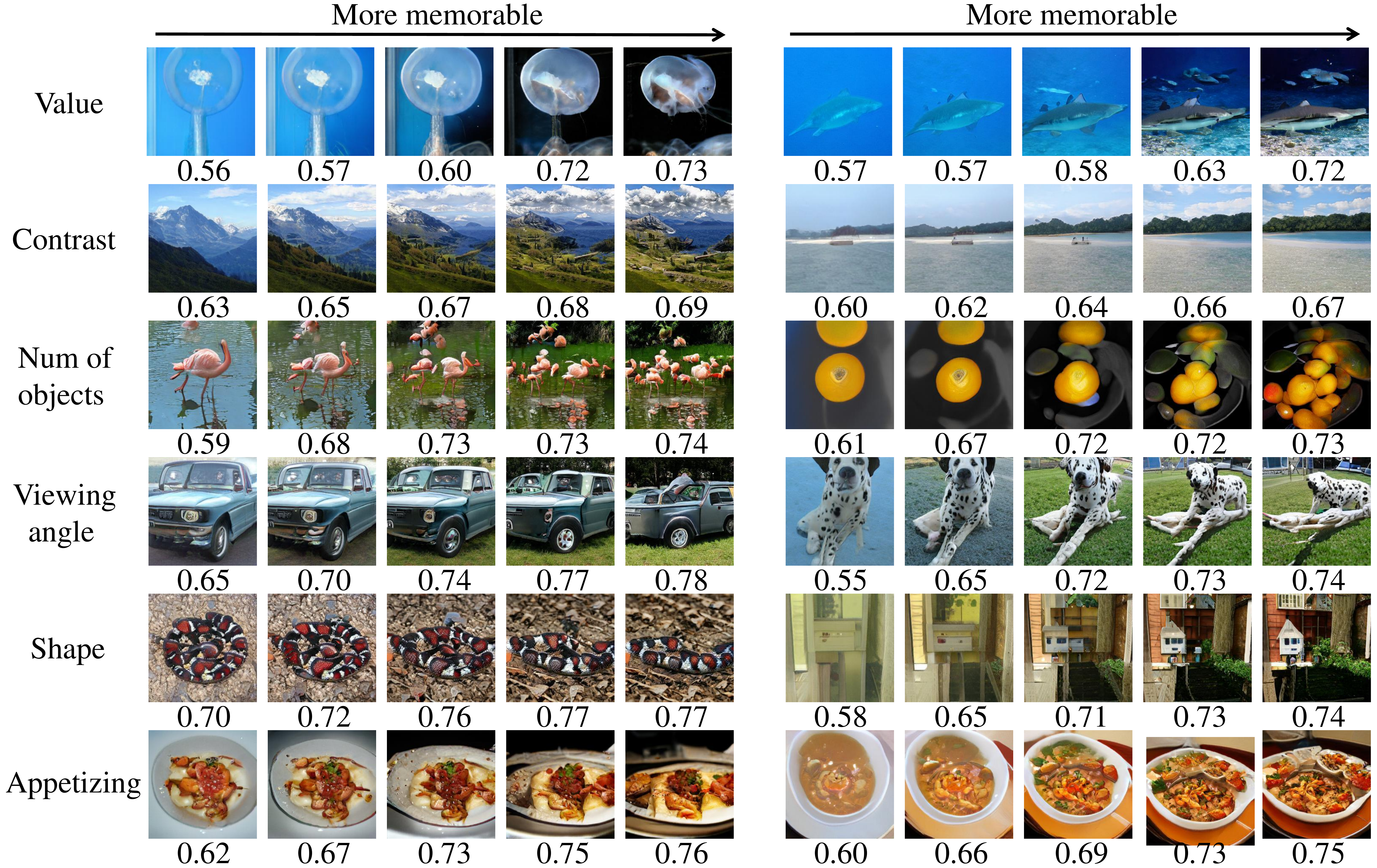}
     \caption{\textbf{Images generated by GANalyze}. We visualize what will happen if we make an image more or less memorable to machines. We summarize 6 trends, where the first 3 of them (value, contrast, and number of objects) are previously found in the quantitative analysis part. We show the results of GANalyze as further confirmations. For certain objects, viewing angle, shape, and appetizing are hidden trends that are unveiled by GANalyze. An overall trend is that GANalyze is often complexifying images to make them more memorable to machines.
     }
     \label{fig:ganalyze}
\end{figure*}
\begin{figure*}[!ht]
     \centering
     \includegraphics[width = 16cm]
     {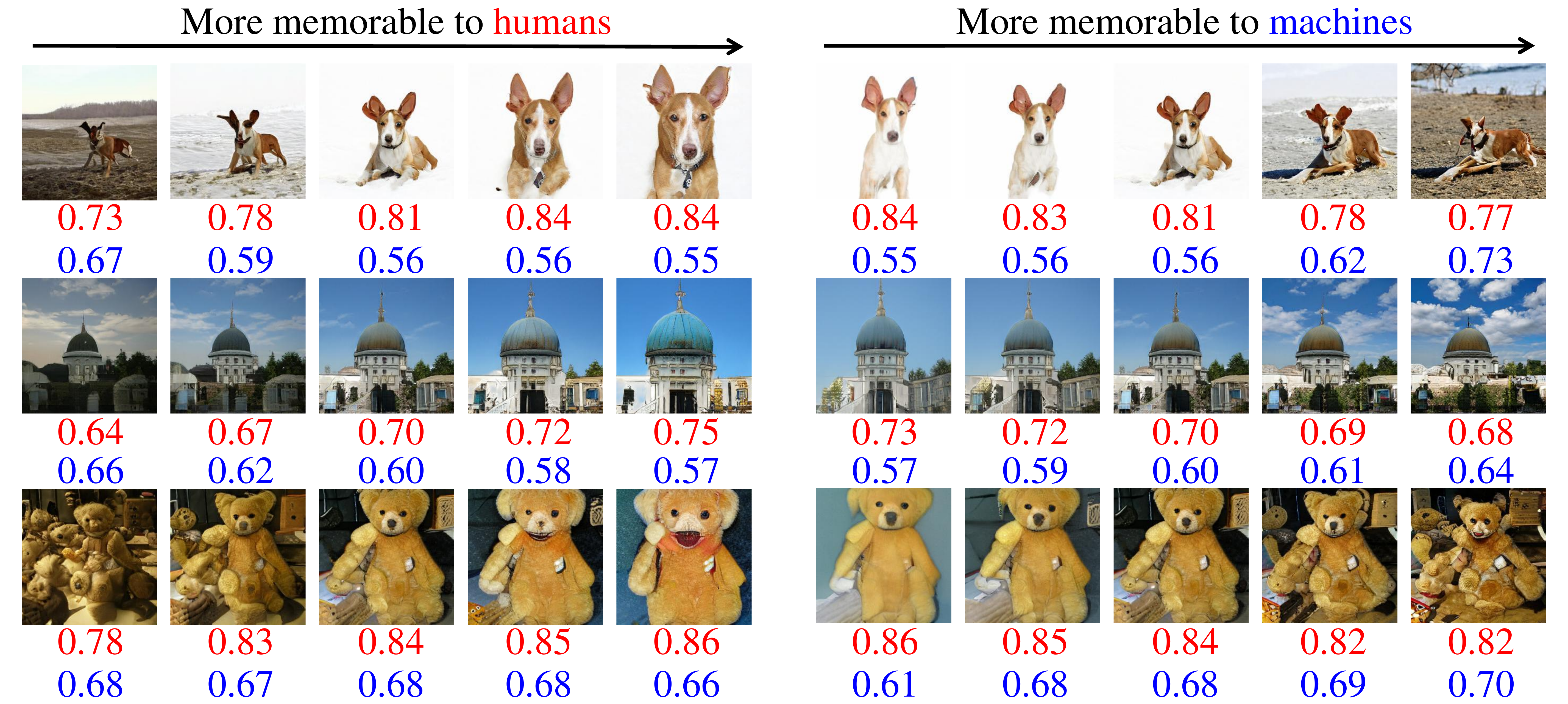}
     \caption{\textbf{A comparison between human memory and machine memory}. We employ GANalyze to make an image more as well as less memorable to both humans and machines. HumanMem scores are labeled in \re{red} while \bl{blue} indicates MahinceMem scores. Generally speaking, simple images are more memorable to humans while complex images are more memorable to machines. 
     }
     \label{fig:ganalyze_compare}
\end{figure*}

\begin{figure*}[!]
     \centering
     \includegraphics[width = 17cm]
     {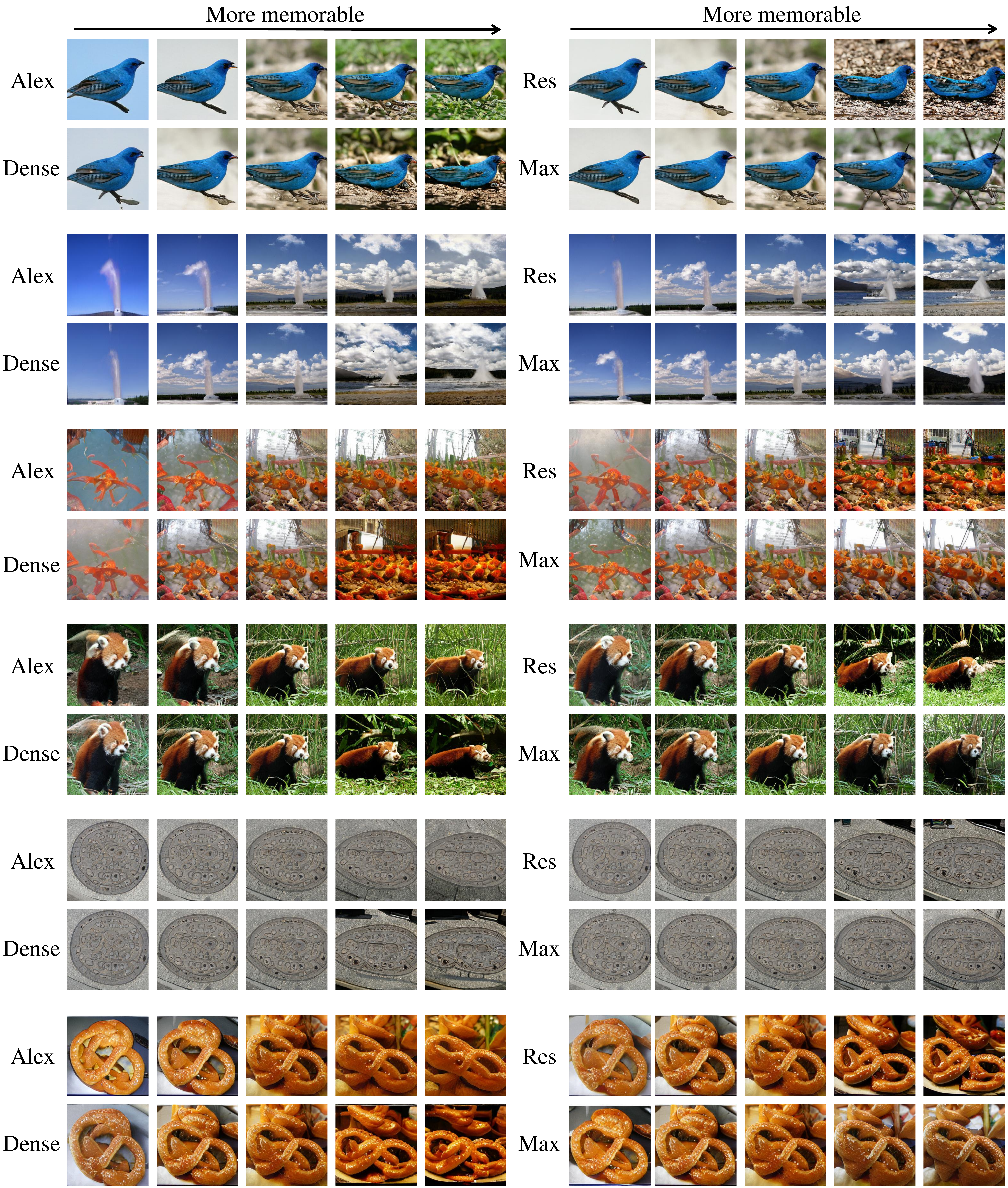}
     \caption{\textbf{A comparison between multiple machines using GANalyze}. Alex, Res, Dense, and Max are abbreviations representing AlexNet, ResNet-50, DenseNet, and MaxViT respectively. Each pair of rows represents a distinct trend, in the following order from top to bottom: value, contrast, number of objects, viewing angle, shape, and appetizing. We continue to discern patterns that recur across diverse machine models. For instance, images securing higher MachineMem scores typically exhibit lower value and strong contrast. Most trends identified within the ResNet-50 translate to other machines.
     }
     \label{fig:ganalyze_multiple_machines}
\end{figure*}

\subsection{GANalyze}
\label{sec:ganalyze}
While the relationship between MachineMem scores and various image attributes has been established, certain concealed factors that potentially enhance an image's memorability for machines remain elusive. Therefore, we leverage the potent capabilities of GANalyze to uncover these hidden elements that could significantly influence MachineMem scores. More specifically, we employ the MachineMem predictor as the Assessor within the GANalyze. This strategy guides the model to manipulate the latent space, increasing or reducing an image's machine memorability. The results of this investigation are depicted in Figure~\ref{fig:ganalyze}.

In the process, we also utilize GANalyze to provide additional validation for the correlations between MachineMem scores and image attributes. We elucidate three observable trends: Value, Contrast, and Number of Objects, as shown from quantitative analysis. In terms of concealed trends, we identify three standout candidates. The newly unearthed trends are summarized below:

$\bullet$ Viewing angles that yield more information, such as side views, are typically more memorable than angles that provide less information.

$\bullet$ Objects that adhere to standard shapes, such as squares or circles, appear less memorable to machines. In contrast, objects with irregular shapes tend to be more memorable.

$\bullet$ Regarding food-related objects, images with high MachineMem scores frequently appear more appetizing.

A critical overarching trend observed across almost all images is the machine tend to memorize complex images. Factors such as contrast, number of objects, viewing angle, shape, and appeal to taste can all be considered manifestations of this pervasive trend.



 


\subsection{Human memory vs.\ machine memory}
As presented in Figure~\ref{fig:lamem}, MachineMem scores and HumanMem scores are very weakly correlated ($\rho$ = $-0.06$). But in GANalyze, which is good at showing global trends, we find machines tend to memorize more complex images, which is on the reverse side of humans that are usually better at memorializing simple images. Such results are presented in Figure~\ref{fig:ganalyze_compare}. Other than ResNet-50, we also explore the correlations between multiple machines (10 other different machines that will be presented in the next section) and humans, however, none of these machines show clear correlations ($\lvert \rho \rvert  \geq 0.15$) with humans. This further suggests MachineMem is very distinct from HumanMem.

\subsection{What images are more memorable to other machines?}
We further incorporate three additional models including AlexNet~\cite{alexnet}, DenseNet121~\cite{huang2019convolutional}, and MaxViT-T~\cite{tu2022maxvit} to visualize the types of images they find more memorable. As depicted in Figure~\ref{fig:ganalyze_multiple_machines}, we employ GANalyze to facilitate a comparative visual representation across multiple machine models. Factors such as value, contrast, viewing angle, appetizing, and complexity continue to serve as reliable predictors of MachineMem scores across various machine models. However, metrics like the number of objects and shape do not consistently apply to all machines. For instance, MaxViT does not display a preference for images containing a larger number of objects. Despite this, the overarching trend remains, that is, machines generally find more complex images memorable. In the following section, we will present a more thorough quantitative analysis concerning machine memorability across various machine models.

\section{Understanding Machine Memory}
HumanMem is considered an intrinsic and stable attribute of an image, consistently recognized by individuals despite disparate backgrounds~\cite{isola2013makes}. This implies that even though human experiences may vary considerably, there remains a commonality in the way humans remember visual data. But does this principle extend to machines?

To enhance our understanding of machine memory, this section seeks to investigate three pivotal questions: Will MachineMem scores keep consistent across different machines? What is the role of varied pre-training knowledge? Is MachineMem Consistent Across Training Settings?

\subsection{Memory across machines}
\label{sec:acrossmachine}
 \begin{figure}[!htb]
     \centering
     \includegraphics[width = 9cm]
     {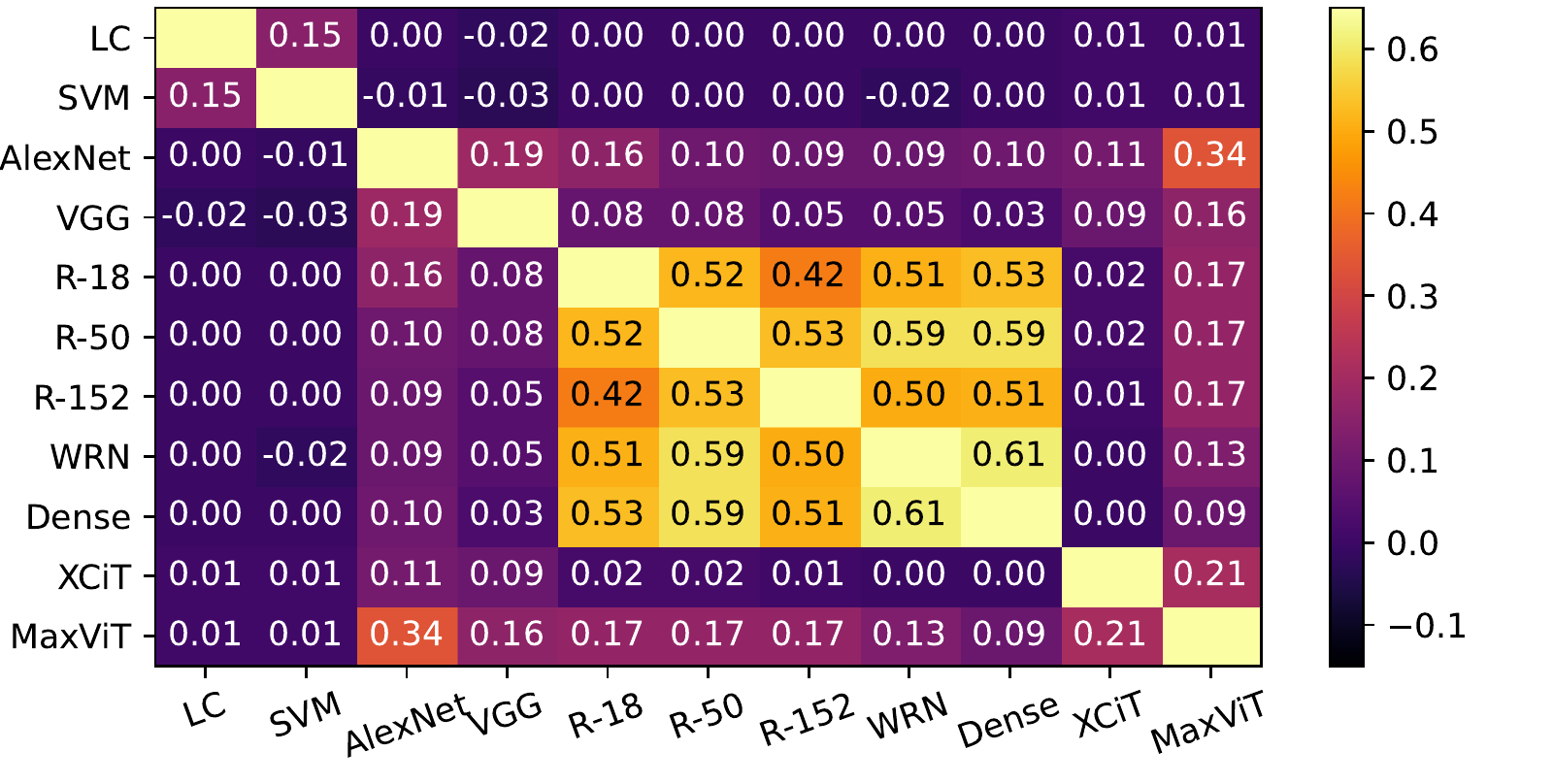}
     \caption{\textbf{Memory across machines}. Each off-diagonal corresponds to Spearman's correlation ($\rho$) of two machines.  Machines within each category are usually strongly correlated, but this trend does not scale to machines across categories. 
     }
     \label{fig:machine}
\end{figure}
We scrutinize the relationships among 11 distinct machines, grouping them into four categories: conventional machines (linear classifier and SVM~\cite{cowan2001magical}), classic CNNs (AlexNet~\cite{alexnet}, VGG~\cite{vgg}), modern CNNs (ResNet-18, ResNet-50~\cite{resnet}, ResNet-152, WRN-50-2~\cite{Zagoruyko2016WRN}, and DenseNet121~\cite{huang2019convolutional}), and recent Vision Transformers (ViTs) (XCiT-T~\cite{ali2021xcit} and MaxViT-T~\cite{tu2022maxvit}). We examine and evaluate the MachineMem scores of 10000 LaMem images as produced by these varying machines. Due to the inherent constraints of conventional machines, we employ a binary classification task (0 \degree and 90 \degree comprising one class, and 180 \degree and 270 \degree forming the other) as their pretext tasks in the initial stage (a) of the MachineMem measurement process. The training parameters are identical for each machine within the same category, although slight variations exist across different categories.

As represented in Figure~\ref{fig:machine}, machines within the same category generally exhibit strong correlations (average $\rho$ of modern CNNs = $0.53$), indicating a tendency to memorize similar images. However, less apparent correlations are observed among machines from distinct categories. For instance, conventional machines, due to their limited modeling capabilities, do not correlate with machines from other categories.

\subsection{Memory across pre-training methods}
\begin{figure}[!htb]
     \centering
     \includegraphics[width = 9cm]
     {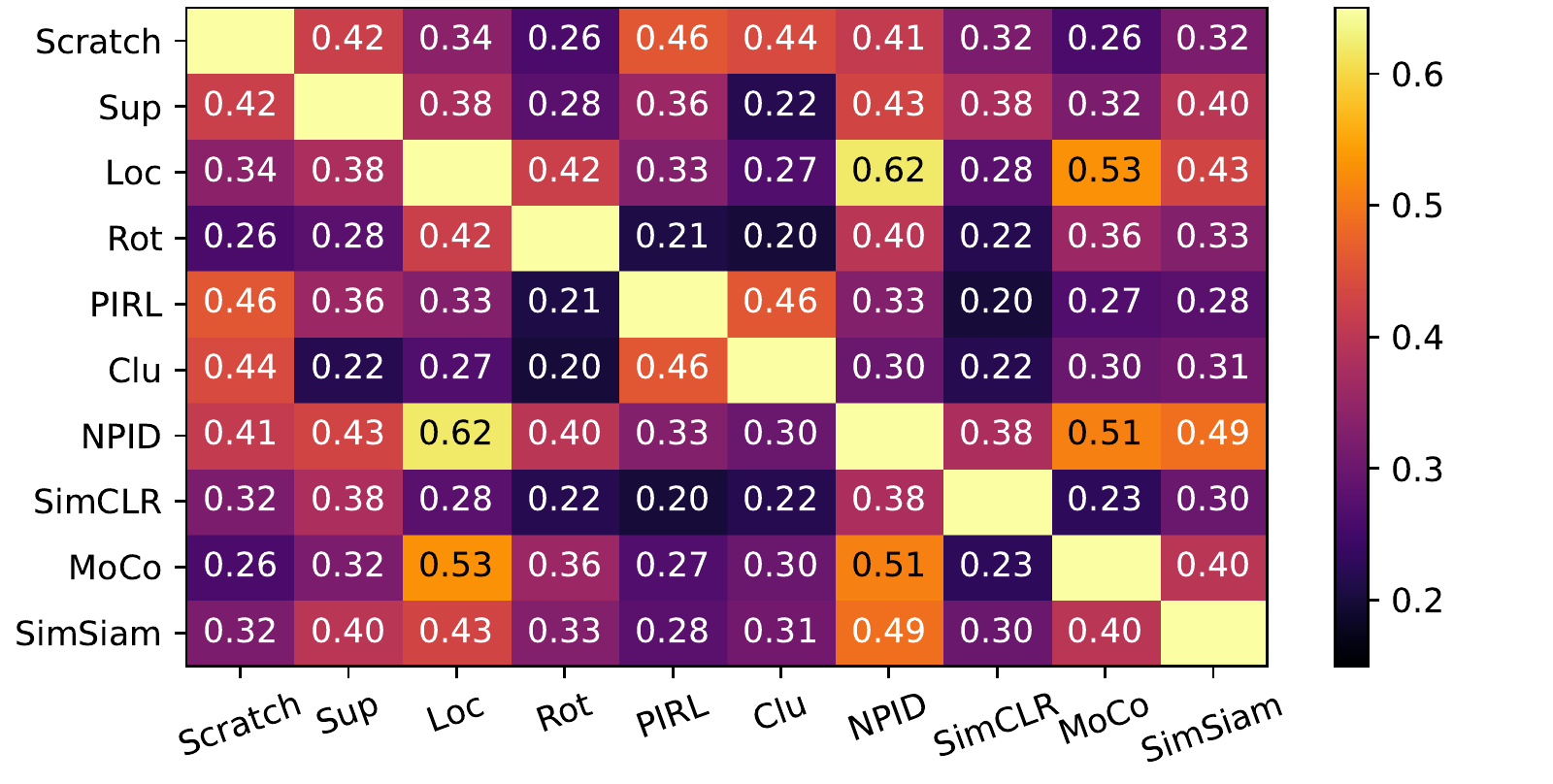}
     \caption{\textbf{Memory across pre-training methods}. Spearman's correlation ($\rho$) of two pre-training methods is presented at each off-diagonal. Though having different prior knowledge, an identical structured machine tends to memorize similar images. 
     }
     \label{fig:pretrain}
\end{figure}
We investigate 9 pre-training methods applicable to a ResNet-50 model. We examine supervised ImageNet classification pre-training and 8 unsupervised methods, which include relative location~\cite{doersch2015unsupervised}, rotation prediction~\cite{gidaris2018rotation}, PIRL~\cite{pirl}, DeepCluster-v2~\cite{deepclu}, and four instance discrimination approaches (NPID~\cite{wu2018unsupervised}, SimCLR~\cite{chen2020simple}, MoCo v2~\cite{chen2020improved}, and SimSiam~\cite{chen2021exploring}). The analysis is carried out on 10000 LaMem images.

The findings are encapsulated in Figure~\ref{fig:pretrain}. The strongest correlation in MachineMem scores ($\rho$ = $0.62$) is observed between the results produced by location prediction and NPID. The weakest correlation ($\rho$ = $0.20$) emerges between PIRL and SimCLR. In general, the memory capabilities of a ResNet-50 model are not significantly influenced by its pre-training knowledge (average $\rho$ = $0.35$). This observation suggests that MachineMem, akin to HumanMem, can be considered an intrinsic and stable attribute of an image, shared across different models despite variations in their pre-training knowledge.

\subsection{Is MachineMem consistent across training settings?}

Human memory remains consistent over time~\cite{isola2013makes}. In a visual memory game, an image that is notably memorable after a few intervening images retains its memorability even after thousands of intervening images. We evaluate a ResNet-50 model across different training settings to discern if MachineMem shares this consistency over varying training configurations.

\paragraph{Number of samples.}
By default, we select the image set size $n$ as 500. We further test this setting with sizes of 50, 1000, and 5000. The Spearman's correlation between the default setting $n = 500$ and these variations are $\rho$ = $0.05$, $\rho$ = $0.66$, and $\rho$ = $0.43$ respectively. A very small $n$ is insufficient to train a stable model and thus fails to accurately reflect memory characteristics. As $n$ increases, the correlations between different $n$ values become increasingly strong.

\paragraph{Number of epochs in stage (a).}
The default number of epochs in stage (a) is set to 60. We test four additional settings: 15, 30, 45, and 75. The Spearman's correlation between the default setting and these variants are $\rho$ = $0.21$, $\rho$ = $0.42$, $\rho$ = $0.55$, and $\rho$ = $0.57$ respectively. The correlation becomes stronger as the number of epochs in stage (a) increases. Once the model undergoes sufficient training (30 epochs), its memory  characteristics stabilizes over multiple epochs in stage (a).

\paragraph{Number of epochs in stage (b).}
We default the number of epochs in stage (b) to 10. Four other settings (1, 4, 7, and 13) are tested. The Spearman's correlation between the default setting and these variants are $\rho$ = $0.31$, $\rho$ = $0.54$, $\rho$ = $0.59$, and $\rho$ = $0.57$. As with human memory, machine memory also appears consistent over time/delay. This finding suggests that for both humans and machines, the correlation between short-term and long-term memory remains strong in such rank memorability measurements.

\section{Implications for computer vision community and potential applications.}

Identifying visually memorable data can lead to practical applications in areas such as data augmentation, continual learning, and generalization. For example, we might design a new data augmentation strategy that can make data more memorable to machines to assist the training of neural networks. In the context of continual learning, it may be advantageous to give greater attention to data that is less memorable.

The way artificial intelligence operates is still vastly different from natural intelligence and creating machines that mimic human behavior remains
challenging. A deeper comprehension of how pattern recognition machines work can facilitate the development of more intelligent machines.

\section{Conclusion}
We propose and study a property of images, \ie{}, machine memorability. Machine memorability shows a cognitive property of machines and can serve as a pathway to help us to further explore machine intelligence.
We hope our findings could provide insights into fundamental advances in computer vision, machine learning, natural language processing, and general artificial intelligence.

\ifCLASSOPTIONcaptionsoff
  \newpage
\fi

\bibliographystyle{IEEEtran}
\bibliography{egbib}









\end{document}